\documentclass[10pt,draftcls,onecolumn]{IEEEtran}

\usepackage[cmex10]{amsmath}
\usepackage{cases}
\usepackage{amsfonts,mathtools}
\usepackage{bm}
\usepackage{cite}
\usepackage{color}
\usepackage{graphicx}	
\usepackage{amssymb,url}
\usepackage{array}	
\usepackage{dsfont}
\usepackage{pgfplots,stfloats}
\usepackage[T1]{fontenc}                 
\usepackage[normalem]{ulem}
\usetikzlibrary{positioning}

\newtheorem{theorem}{Theorem}
\newtheorem{proposition}{Proposition}
\newtheorem{definition}{Definition} 
\newtheorem{lemma}{Lemma}
\newtheorem{assumptions}{Assumptions}
\newtheorem{corollary}{Corollary}
\newtheorem{remark}{\emph{Remark}}    

\usepackage{cancel}
\usepackage{float}

\usepackage{float}
\usepackage{stmaryrd}
\newcommand{\mkv}{-\!\!\!\!\minuso\!\!\!\!-}
\usepackage{subcaption}

\ifCLASSINFOpdf
\else
\fi
\begin{document}
%

\title{The Role of Information Complexity and\\[-3mm]  Randomization in Representation Learning}

%
%
%
%

\author{Matias Vera,~\IEEEmembership{Student~Member,~IEEE,}
        Pablo~Piantanida,~\IEEEmembership{Senior~Member,~IEEE,}
        and~Leonardo~Rey~Vega,~\IEEEmembership{Member,~IEEE}
\IEEEcompsocitemizethanks{\IEEEcompsocthanksitem M. Vera is with the Facultad de Ingenier\'ia, Universidad de Buenos Aires (FIUBA), Argentina. Email: mvera@fi.uba.ar.}
 
 \IEEEcompsocitemizethanks{
\IEEEcompsocthanksitem  P. Piantanida is with Laboratoire des Signaux et Syst\`emes (L2S, UMR8506), CentraleSup\'elec-CNRS-Universit\'e Paris-Sud, Gif-sur-Yvette, France. Email: pablo.piantanida@centralesupelec.fr.}

\IEEEcompsocitemizethanks{\IEEEcompsocthanksitem L.~Rey~Vega is with the Dept. of Electronics (FIUBA) and CSC-CONICET, Buenos Aires, Argentina. Email: lrey@fi.uba.ar.}

}

\IEEEtitleabstractindextext{%
\begin{abstract}
A grand challenge in representation learning is to learn the different explanatory factors of variation behind the high dimen- sional data. Encoder models are often determined to optimize performance on training data when the real objective is to generalize well to unseen data. Although there is enough numerical evidence suggesting that noise injection (during training) at the representation level might improve the generalization ability of encoders, an information-theoretic understanding of this principle remains elusive. This paper presents  a sample-dependent bound on the generalization gap of the cross-entropy loss that scales with the  \emph{information complexity} (IC) of the representations, meaning  the mutual information  between inputs and their representations. The IC is empirically investigated for standard multi-layer neural networks with SGD on MNIST and CIFAR-10 datasets; the behaviour of the gap and the IC appear to be in direct correlation, suggesting that SGD selects encoders to implicitly minimize the IC. We specialize the IC to study the role of Dropout on the generalization capacity of deep encoders which is shown to be directly related to the \emph{encoder capacity}, being a measure of the distinguishability among samples from their representations. Our results support some recent regularization methods.
\end{abstract}

}
\maketitle


\IEEEdisplaynontitleabstractindextext

%
\IEEEpeerreviewmaketitle
\allowdisplaybreaks

\section{Introduction}

Deep neural networks has turned out to be very good at discovering structures in high-dimensional data and has dramatically improved the state of the art~\cite{LeCun2015Deep}. The global learning task is decomposed into a hierarchy of layers with nonlinear processing,  having great success not only due to their ability to fit any type of dataset but also to generalize incredible well.  The representational capabilities of neural networks~\cite{Bengio2013Representation} have drawn significant interest from the machine learning community, starting from universal approximation of mesurable Borel functions~\cite{Hornik1991251,Cybenko1989,Mhaskar1993,NIPS2011_4350}, sample complexity results~\cite{661502}, and more recently~\cite{DBLP:journals/corr/EldanS15,DBLP:journals/corr/MhaskarP16, DBLP:journals/corr/Telgarsky16, DBLP:journals/corr/CohenS16} have established  benefits of approximating with depth networks. Although these works address the problem of determining which functions can be well approximated by certain families of neural networks, they do not consider the representational ability of randomized encoders. Although numerical evidence supports  that the depth of architecture can be very important for statistical efficiency,  the actual goal of learning is neither accurate estimation of model parameters~\cite{Barron1994} nor compact representation of the data itself~\cite{Rissanen1978Modeling,86996}; rather, we are mostly interested in the generalization capabilities, meaning its ability to successfully apply rules extracted from previously seen data to characterize unseen data. 

According to the statistical learning theory~\cite{boucheron_bousquet_lugosi_2005}, models with many parameters tend to \emph{overfit} by representing the learned data too accurately, therefore diminishing their ability to generalize to unseen data. In order to reduce this `generalization gap', i.e., the difference between `training error' and `test error',  conventional regularization methods~\cite{vapnik98statlearn} include weight penalty based methods (usually through the addition of a penalty term that is  closely related to weight decay~\cite{NIPS2012_4824}), averaging over multiple models, where each member of the ensemble is trained with a different subsample of the input data~\cite{Breiman1996}, or artificial data augmentation. A different training method based on noise injection to penalize complex models indirectly by adding noise to weight connections and weight changes of multilayer feedforward neural networks was investigated in~\cite{105415, 155944, 809073, 170429,  5674088, Sietsma199167}. References~\cite{227178,doi:10.1162/neco.1996.8.3.643, doi:10.1162/neco.1995.7.1.108, doi:10.1162/neco.1997.9.5.1093, 5674088} showed the objective that results from injecting noise during backpropagation training consists of two terms: the conventional training set error while the other is a Tikhonov regularization term which improves the generalization ability. 

Another more recent breakthrough has been the development of \emph{Dropout}~\cite{srivastava2014dropout} for training deep neural networks which consists in randomly dropping units during training to prevent their co-adaptation. This concept has been previously introduced in the context of denoising auto-encoders (DAEs)~\cite{Vincent2010Stacked} where noise is added to the input units of an autoencoder and the network is trained to reconstruct the noise-free input. In~\cite{Alain2014What} has been shown that  when a DAE (or a contractive AE) is trained with small Gaussian noise and squared error loss, it estimates a score proportional to the difference between the input and its reconstruction. As a matter of fact, experimental results suggest that the use of noise induces  major improvements in the misclassification probability. Recent works~\cite{2016arXiv161101353A, DBLP:journals/corr/AlemiFD016} explore an information-based generalization of Dropout  inspired from the heuristic \emph{Information Bottleneck method}~\cite{Tishby1999information}, yielding a slightly more general form of the variational auto-encoder~\cite{kingma2013auto}. Despite of numerical evidence or some existent explanations, a lot of work needs to be done if theory has to meet practice. 

\subsection{Focus and Related Work}

In this work, we study the underlying information-theoretic principles of the effective generalization capacity in representation (encoder) models under the cross-entropy loss. In particular, we aim at investigating the following general question: 
\emph{What is the effect of using randomized encoders on the generalization gap of the learned representations?} More specifically:  \emph{can we bound the generalization gap in terms of information measures, involving a randomized encoder (noise distribution), the inputs and their corresponding representations?}  Despite the fact that the concepts of information and noise are natural to the problem of lossy data compression~\cite{Shannon1959Coding}, the benefits (or even necessity) of limiting the amount of information (with noise) in representations remains elusive.  

Recently, there has been much interest in understanding the importance of implicit regularization. Numerical experiments in~\cite{DBLP:journals/corr/NeyshaburTS14,DBLP:journals/corr/NeyshaburTS15} demonstrate that network size may not be the main form of capacity control and hence, some other unknown form plays a central role in learning multi-layer feed-forward networks. Further work in~\cite{DBLP:journals/corr/ZhangBHRV16} found that, in contrast with classical convex empirical risk minimization,  regularization plays a rather different role in deep  learning. From the theoretical viewpoint regularization seems to be an indispensable component while convincing experiments support the idea that the absence of all regularization does not necessarily induces poor generalization. 
This question was also approached via the rate-distortion function~\cite{Shamir:2010:LGI:1808343.1808503,DBLP:journals/corr/Shwartz-ZivT17} by using an intuitive analogy between the IB method and deep models~\cite{DBLP:journals/corr/TishbyZ15}. Along the same line of work, \cite{Russo_Zou_2015,Xu_Raginsky_2017} have derived interesting results proving that the square-root of the mutual information between the training inputs and the inferred parameters provides a concise bound on the gap, which crucially depends on mapping of the training set into the network parameters, whose characterization is not an easy task. \cite{Achille_Soatto_2017} explored how the use of an IB  objective on the network parameters (and not on the representations) may help to avoid overfitting while enforcing invariant representations. On the other side, direct use of statistical learning theory such as \emph{Rademacher complexity}~\cite{Bartlett2001}, \emph{VC-dimention}~\cite{vapnik98statlearn} and \emph{uniform stability}~\cite{Bousquet:2002:SG:944790.944801}  seem to be inadequate to explain some of the unexpected observations obtained from numerical experimentation. 

\subsection{Contributions}

Our goal here is different: we are interested in investigating whether there may be a different form of information-theoretic (perhaps implicit) regularization that controls the generalization gap and would like to study whether representation learning benefits from training with  randomized (noisy) encoders. Toward this goal, the models considered here are allowed to work on rather general encoder classes but restricted to  finite input and label alphabets only. The two central contributions are as follows.

\begin{itemize}
\item Theorem~\ref{thm:regularizar} will give the rigorous statement of the generalization gap bound that is the basis of this work. In contrast to standard approaches based on \emph{VC-dimention} or \emph{covering-numbers}, this sample-dependent bound: (a) scales with $\log(n)/\sqrt{n}$ and $1/\sqrt{n}$ w.r.t. the length $n$ of training samples; (b) has dependence on the  \emph{information complexity} (IC), i.e., mutual information between the inputs and their representations,  against a reference selected encoder and on a measure of the decoder efficiency  (with explicit dependence on the selected decoder); (c) has no dependence on combinatorial parameters; (d) provides an information-theoretic interpretation of the IC  based on the notion of encoder capacity and its connection to existent regularization methods (see Section~\ref{sec-info-complex}); (e) allows to obtain, in some cases, explicit expressions of the IC bounds (see Section~\ref{Section-4}). Beside our results may not lead to the tightest bounds, they are intended to reflect the importance of information-theoretic concepts in the problem of representation learning and the different trade-offs that can be established between information measures and quantities of interest in statistical learning theory. 

\item An empirical investigation  of the interplay between generalization and information complexity in deep neural networks on the standard datasets MNIST and CIFAR-10 is presented in Section~\ref{Section-6}. Our simulations study the ability of the  IC to predict  the behaviour of the generalization capacity in front of several factors:  statistical variations of data during train, amount of noise in representations, and  network size.  This study illuminates some interesting  features of the IC: (a) it is an increasing function of the network (hidden units) size; (b) it evolves similarly to the gap with variations of noise; and (c)  it detects well the mismatch (or inadequacy) of data during the training phase, which consists of copying the data where the true labels were replaced by random or roll labels.
\end{itemize}

The outline of this paper is as follows. In Section~\ref{Section-2}, we present necessary definitions and the problem statement. In Section~\ref{Section-3}, we present our upper bound on generalization gap and an information-theoretic interpretation of the information complexity together with its connection to existent regularization methods, while proofs are provided in Section~\ref{sec:proof_main_theorem} and Appendices.  Section~\ref{Section-4} provides applications and bounds for specific models on the information complexity. Section~\ref{Section-6} provides experimental results. Conclusions are presented in Section~\ref{Section-7}.

\subsection*{Notation and conventions}

We use upper-case letters to denote random variables and lower-case letters to denote realizations of random variables. In some cases we will need to emphasize that a given random variable is vector-value one. In these cases, with upper-case bold letters we indicate random vectors and with lower-case bold letters we denote realizations of random vectors or constant vectors (the right interpretation should be clear from context).  As usual, $\mathbb{R}$ is the notation for the real numbers and $\mathbb{N}$ for natural ones. Superscripts are used to denote the length of the vectors and subscripts denote the index of each entry in the vector. The notation $|\cdot|$ will be used interchangeably for the usual absolute value of a real number and for the cardinality of a set. Its correct use should be clear from the context. Let $\mathcal{P}(\mathcal{X})$ denote the set of all probability measures in the underlying alphabet $\mathcal{X}$. The probability mass function (pmf) of random variable $X$ is denoted generically as $P_{X}$ (on occasions we will also write $X\sim P_X$). With $P_X(x)$ we indicate the value of the pmf at $x\in\mathcal{X}$. Similarly conditional pmf of random variable $Y$ with respect to $X$ is denoted as $P_{Y|X}$ and with $P_{Y|X}(y|x)$ we denote the value it takes at $(x,y)\in\mathcal{X}\times\mathcal{Y}$. With $X\mkv Y\mkv Z$ we denote a Markov chain, that is ${P}_{X|YZ}= {P}_{X|Y}$.  All alphabets are assumed to be finite, i.e., with an arbitrarily large but  finite number of atoms. $ \mathbb{E}_{P}[\cdot]$ denotes the mathematical expectation w.r.t. the probability distribution $P$. Empirical distributions based on data samples are denoted by $\hat{P}_{X}$. In some situations we will need to consider the full vector of length $|\mathcal{X}|$ that contains the values of distribution $P_X$. We will denote these quantities as $\mathbf{P}_X$. In other cases we will need to write all the values of an arbitrary function $f:\mathcal{X}\rightarrow\mathbb{R}$ as a $|\mathcal{X}|$-length vector. This will be indicated as $\left\{f(x)\right\}_{x\in\mathcal{X}}$. With $\|\cdot\|_2$ we denote the usual Euclidean norm of a vector and with $\langle\cdot,\cdot\rangle$ the canonical inner product. With $\mathds{1}\left[\cdot\right]$ we will denote the usual indicator function. For an arbitrary differentiable function $f:\mathbb{R}^{L}\rightarrow\mathbb{R}$, we will denote its gradient as $\frac{\partial f(\mathbf{x})}{\partial \mathbf{x}}$. Asymptotics will be important in our analysis. We will say that $f(x)=\mathcal{O}\left(g(x)\right)$ when $x\rightarrow x_0$ if there exists a vicinity of $x_0$ and $M>0$ such that $|f(x)|\leq M |g(x)|$ for all $x$ in such vicinity. Similary, we will say that $f(x)=o\left(g(x)\right)$ when $x\rightarrow x_0$ if for every $\epsilon>0$ there exists a vicinity of $x_0$ where $f(x)\leq \epsilon|g(x)|$ for all $x$ in such vicinity. All logarithms are taken with base $e$. The involved information measures in our analysis will be~\cite{cover}: the \emph{entropy}  $ \mathcal{H}(P_X)\coloneqq \mathbb{E}_{P_X}\left[-\log { {P}_X}(X)\right]$; the \emph{conditional  entropy} $ \mathcal{H}(P_{Y|X}|P_X)\coloneqq \mathbb{E}_{P_{X}}\mathbb{E}_{P_{Y|X}}\left[-\log {P}_{Y|X} (Y|X)\right]$; the \emph{relative entropy}: 
\begin{equation}
\mathcal{D}( {P}_{X}\| {Q}_{X})\coloneqq  \left\{  
\begin{array}{lll}
\displaystyle \mathbb{E}_{ P_{X}}\left[\log \frac{ {P}_{X}(X)}{ {Q}_X(X)}\right] & & \textrm{if $P_X \ll Q_X$}\\
+\infty & & \textrm{otherwise,}
\end{array}
\right.
\end{equation}
where we use notation $P_X \ll Q_X$ to denote that the probability measure $P_X$ is \emph{absolutely continuous} with respect to $Q_X$. The conditional  \emph{relative entropy} is defined by 
$
\mathcal{D}( {P}_{Y|X}\| {Q}_{Y|X}| {P}_{X})\coloneqq \mathbb{E}_{ {P_X}}\left[ \mathcal{D}\big( {P}_{Y|X}(\cdot| X)\| {Q}_{Y|X}(\cdot| X)\big) \right] 
$
and the \emph{mutual information} is given by $\mathcal{I}(P_X;{P}_{Y|X})\coloneqq \mathcal{D}( {P}_{Y|X} \| {{P}_Y} | P_X)$.

\section{Representation and Statistical Learning}\label{Section-2}

In this section, we introduce the framework in which leaning from examples is to be studied. We develop precise notions of risk and its generalization gap, and discuss the information-theoretic factors upon which these depend.

\subsection{Basic Definitions}

We are concerned with the problem of pattern classification which is about guessing (or predicting) the unknown class of an observation. An observation is often a sample $x\in \mathcal{X}$ presented to the learner about a target concept $y\in \mathcal{Y}$ (the concept class). In our model we simple assume $(\mathcal{X}, \mathcal{Y})$ are abstract discrete spaces equipped with a $\sigma$-algebra. In classification, one creates a function $h:\mathcal{X}\rightarrow \mathcal{Y}$ which represents one's guess of $y$ given $x$. 

\begin{definition}[Probability of misclassification] An $| \mathcal{Y}|$-ary classifier is defined by a (possibly stochastic) decision rule $Q_{\hat{Y}|X}:\mathcal{X} \rightarrow \mathcal{P}(\mathcal{Y})$, where $\hat{Y}\in\mathcal{Y}$ denotes the random variable associated to the classifier output and $X$ is the information source. The probability of misclassification of a decision rule $Q_{\hat{Y}|X}$  with respect to a data distribution $P_{XY}$ is given by
\begin{equation}
P_{\mathcal{E}}\big(Q_{\hat{Y}|X}\big) \coloneqq 1- \mathbb{E}_{ P_{XY}}\left[ Q_{\hat{Y}|X}(Y|X)\right]. \label{eq-true-error}
\end{equation}
\end{definition}

Minimizing over all possible conditional classifiers  $Q_{\hat{Y}|X}$ gives the smallest average probability of misclassification and an optimum classifier chooses the hypothesis $\hat{y}\in\mathcal{Y}$ with largest posterior probability $P_{Y|X}$ given the observation $x$, that is the \emph{Maximum a Posteriori (MAP)} decision. The MAP test that breaks ties randomly with equal probability is given by\footnote{In general, the optimum classifier given in (\ref{eq:optimum_map}) is not unique. Any conditional pmf with support in $\mathcal{B}(x)$ for each $x\in\mathcal{X}$ will be equally good.}
\begin{equation}
Q_{\hat{Y}|X}^{\textrm{MAP}} (y|x)\coloneqq 
\left\{ 
\begin{array}{lll}
\displaystyle \frac{1}{|\mathcal{B}(x)|}, & & \textrm{if $y\in \mathcal{B}(x)$}\\
0, & & \textrm{otherwise}
\end{array}
\right.\label{eq:optimum_map}
\end{equation}
where the set $\mathcal{B}(x)$ is defined as: 
$
\mathcal{B}(x)\coloneqq\left\{ y\in \mathcal{Y}\,\big | \,P_{Y|X}(y|x) = \max\limits_{y^\prime\in \mathcal{Y}} P_{Y|X}(y^\prime|x)  \right\}.
$
This classification rule is called \emph{Bayes decision rule}. Bayes decision rule is optimal in the sense that no other decision rule has a smaller probability of misclassification. It is straightforward to obtain the following lemma:

\begin{lemma}[Bayes error] 
The misclassification error rate of the Bayes decision rule is given by
\begin{eqnarray}
P_{\mathcal{E}}\big(Q_{\hat{Y}|X}^{\textrm{MAP}}\big)   = 1- \mathbb{E}_{P_X} \left[ \max\limits_{y^\prime\in \mathcal{Y}}  P_{Y|X} (y^\prime|X)\right]. 
\end{eqnarray}
\end{lemma}

Nevertheless, finding the Bayes decision rule requires knowledge of the underlying distribution $P_{XY}$, but typically in applications these distributions are not known. In fact, even a parametric form or an approximation to the true distribution is unknown. In this case, the learner tries to overcome this lack of knowledge by resorting to labeled examples. In addition, the probability of misclassification using the labeled examples has the particularity that it is mathematically hard to solve for the optimal decision rule. As a consequence, it is common to work with a surrogate (information measure) function given by the \emph{logarithmic loss}:


\begin{lemma}[Surrogate based on the average logarithmic loss]\label{lemma-surogate}
A natural surrogate for the probability of misclassification $P_{\mathcal{E}}(Q_{\hat{Y}|X})$ corresponding to a classifier $Q_{\hat{Y}|X}$ is given by the average logarithmic loss defined by $\mathbb{E}_{{P}_{XY}}\left[ -\log Q_{\hat{Y}|X}(Y|X)\right]$ and which satisfies: 
\begin{equation}
P_{\mathcal{E}}( Q_{\hat{Y}|X} ) \leq 1-\exp\left(-\mathbb{E}_{{P}_{XY}}\left[ -\log Q_{\hat{Y}|X}(Y|X)\right]\right).\label{eq-surrogate-bound}
\end{equation}
A lower-bound to the logarithmic loss can be computed as:
\begin{equation}
\mathbb{E}_{{P}_{XY}}\left[ -\log Q_{\hat{Y}|X}(Y|X)\right] \geq  \mathcal{H}(P_{Y|X}| P_X).
\end{equation}
\end{lemma}
The logarithmic loss can provide an effective and better behaved surrogate for the particular problem of minimizing the probability of misclassification~\cite{Bengio2013Representation}. Evidently, the optimal decision rule for the logarithmic loss is $Q_{\hat{Y}|X} \equiv  P_{Y|X}$. This does not match in general with the optimal decision rule for the probability of misclassification $Q_{\hat{Y}|X}^{\textrm{MAP}}$ in expression~\eqref{eq:optimum_map}. 
Although the logarithmic loss may induce an irreducible gap with respect to the probability of misclassification,  it is clear that when the true $P_{Y|X}$ concentrates around a particular value $y(x)$ for each $x\in\mathcal{X}$  (which is necessary for a statistical model $P_{Y|X}$ to induce low probability of misclassification) this gap could be significantly reduced.

\subsection{Learning Data Representations}
 
We will concern ourselves with learning representation models (randomized encoders) and inference models (randomized  decoders) from examples of these probabilistic mappings. In other words, learning target probability distributions  which are assumed to belong to some class of distributions. The problem of finding a good classifier can be  divided into that of simultaneously finding a (possibly randomized) encoder $Q_{U|X}:\mathcal{X}\rightarrow\mathcal{P}(\mathcal{U})$ that maps raw data to a higher-dimensional (feature) space $\mathcal{U}$ and a soft-decoder $Q_{\hat{Y}|U}:\mathcal{U}\rightarrow\mathcal{P}(\mathcal{Y})$ which maps the representation to a probability distribution on the label space $\mathcal{Y}$. Although  these mappings induce an equivalent classifier: 
 \begin{equation}
Q_{\hat{Y}|X}(y|x) = \sum\limits_{u\in\mathcal{U}}   Q_{U|X} (u|x)Q_{\hat{Y}|U} (y|u) ,  \label{eq-clasifier}
\end{equation}
the computation of the later expression requires marginalizing out $u \in\mathcal{U}$ which is in general computationally hard due to the exponential number of atoms  involved in representations. A variational upper bound is commonly used to rewrite this intractable problem into:
\small
\begin{equation}
\mathbb{E}_{{P}_{XY}}\left[ -\log Q_{\hat{Y}|X}(Y|X)\right] \leq   \mathbb{E}_{{P}_{XY}} \mathbb{E}_{{Q}_{U|X}}  \left[ -\log Q_{\hat{Y}|U}(Y|U)\right], 
\end{equation}\normalsize
which simply follows by applying Jensen inequality~\cite{cover}. This bound induces the well-known \emph{cross-entropy  risk} defined below: 
\begin{definition}[Cross-entropy loss  and risk]\label{def-Logarithmic-loss}
Given two randomized mappings $Q_{U|X}:\mathcal{X}\rightarrow\mathcal{P}(\mathcal{U})$ and $Q_{\hat{Y}|U}:\mathcal{U}\rightarrow\mathcal{P}(\mathcal{Y})$, define the average (over representations) \emph{cross-entropy loss} as:
\begin{align}
\ell\big(\mathbf{Q}_{U|X}(\cdot|x),\mathbf{Q}_{\hat{Y}|U}(y | \cdot)\big)&\coloneqq\big\langle \mathbf{Q}_{U|X}(\cdot|x) , -\log \mathbf{Q}_{\hat{Y}|U}(y | \cdot ) \big\rangle\nonumber\\
&=- \sum\limits_{u\in\mathcal{U}} {Q}_{U|X}(u|x) \log Q_{\hat{Y}|U}(y | u). \label{eq-true-loss}
\end{align}
We measure the expected performance of $({Q}_{U|X},Q_{\hat{Y}|U})$  via the \emph{risk} function: 
\begin{equation}
\hspace{-0.35cm}\mathcal{L}(Q_{\hat{Y}|U},Q_{U|X}) \coloneqq  \mathbb{E}_{{P}_{XY}}   \big[ \ell\big(\mathbf{Q}_{U|X}(\cdot|X),\mathbf{Q}_{\hat{Y}|U}(Y | \cdot)\big)  \big] \label{logarithmic-loss-risk}
\end{equation}
\end{definition}

In the standard methodology for learning deep representations, we consider $K$-th randomized encoders $\{Q_{U_k|U_{k-1}}\}_{k=1}^K$ with $U_0\equiv X$ instead of one randomized encoder $Q_{U|X}$. Although this appears at first to be more general, it can be casted formally using the one-randomized encoder formulation as being induced by the marginal distribution that relates  the input and the output layer of the network. Therefore any result on the one-layer formulation immediately implies a result on the $K$-th layer formulation and thus, we shall thus focus on the one-layer case without loss of generality. 

\begin{lemma}[Optimal randomized decoders]\label{lemma-optimal-decoder}
The minimum cross-entropy loss  risk satisfies:
\begin{IEEEeqnarray}{rCl}
\inf\limits_{Q_{\hat{Y}|U}:\,\mathcal{U}\rightarrow\mathcal{P}(\mathcal{Y})} \mathcal{L}(Q_{\hat{Y}|U},Q_{U|X}) &=&   \mathcal{H}(Q_{Y|U}|Q_U),
  \end{IEEEeqnarray}
where 
\begin{equation}
Q_{{Y}|U}(y|u)=\frac{\sum\limits_{x\in\mathcal{X}}Q_{U|X}(u|x)P_{XY}(x,y)}{\sum\limits_{x\in\mathcal{X}}Q_{U|X}(u|x)P_X(x)}.
\end{equation}
\end{lemma}

\begin{IEEEproof}
It is not difficult to check that: $\mathcal{L}(Q) =  \mathcal{D}\left(Q_{Y|U} \big \|Q_{\hat{Y}|U} \big| Q_U\right) + \mathcal{H}(Q_{Y|U}|Q_U) $ with  $Q\coloneqq(Q_{\hat{Y}|U},Q_{U|X})$. The proof follows by using the positivity of the relative entropy. 
\end{IEEEproof}

In order to connect formally  the probability of misclassification to the corresponding risk associated to the cross-entropy loss introduced in Definition~\ref{def-Logarithmic-loss}, it is convenient to first introduce the rate-distortion function. 

\begin{definition}[Rate-distortion function \cite{cover}] \label{def-rate-distortion}
The rate-distortion (RD) function  of a random variable $Y\in\mathcal{Y}$ with distortion (loss) function $d(y,u)\coloneqq 1- Q_{\hat{Y}|U}(y|u)$ is defined as:
\begin{equation}
\mathcal{R}_{Y,Q_{\hat{Y}|U}}(D) \,\coloneqq\hspace{-0.2cm}\min_{\rule{0mm}{4.3mm}\substack{P_{\hat{U}|Y}:\, \mathcal{Y}\,\rightarrow\mathcal{P}(\mathcal{U}) \\  \mathbb{E}_{P_{\hat{U}Y} }[1- Q_{\hat{Y}|U}(Y|U)] \, \leq\, D}}\hspace{-0.2cm} \mathcal{I}\big(P_Y;P_{\hat{U}|Y}\big) \,.\label{eq-rate-distortion}
\end{equation}
where $P_{\hat{U}Y}=P_{\hat{U}|Y}P_Y$. Furthermore, it is easy to check that there exists $D>0$ such that $\mathcal{R}_{Y,Q_{\hat{Y}|U}}(D)$ is finite and we denote the minimum of such values of $D$ by $D_{\min}$ with $\displaystyle R_{\max}\coloneqq  \lim_{D\to D_{\min}+}\mathcal{R}_{Y,Q_{\hat{Y}|U}}(D)$. 
\end{definition}

Moreover, it is easy to show that $\mathcal{R}_{Y,Q_{\hat{Y}|U}}(D)$ is positive, monotonically decreasing and convex. In addition, we can define the function:
\begin{equation}
\mathcal{R}_{Y,Q_{\hat{Y}|U}}^{-1}(I)\,\coloneqq  \inf\big\{D\in\mathbb{R}_{\geq 0}: \mathcal{R}_{Y,Q_{\hat{Y}|U}}(D)\leq I \big\}.
\label{eq:inverse_rate}
\end{equation}
This is known as the \emph{distortion-rate}\cite{cover} function\footnote{It is worth to mention that by using $\mathcal{R}_{Y,Q_{\hat{Y}|U}}^{-1}(I)$ we are abusing notation. This is because in general is not true  that $\mathcal{R}_{Y,Q_{\hat{Y}|U}}(D)$ is injective for every $D\geq 0$. However, when $I\in[R_{\min},R_{\max})$ with $R_{\min}\coloneqq \mathcal{R}_{Y,Q_{\hat{Y}|U}}(D_{\max}) $ and $D_{\max}\coloneqq \min\limits_{u\in\mathcal{U}} \mathbb{E}_{P_Y} \left[1- Q_{\hat{Y}|U}(Y|u)\right]$, under some very mild conditions on $P_Y$ and $Q_{\hat{Y}|U}(y|u)$, $\mathcal{R}_{Y,Q_{\hat{Y}|U}}^{-1}(I)$ is the true inverse of $\mathcal{R}_{Y,Q_{\hat{Y}|U}}(D)$, which is guaranteed to be injective in the interval $D\in(D_{\min},D_{\max}]$.}. Again, the function $I\mapsto \mathcal{R}_{Y,Q_{\hat{Y}|U}}^{-1}(I)$ is positive and monotonically decreasing. 

The following lemma provides an upper and lower bounds on the probability of misclassification via the distortion-rate function and the cross-entropy loss. 

\begin{lemma}[Probability of misclassification and cross-entropy loss]\label{lemma-surogate-representations}
The probability of misclassification $P_{\mathcal{E}}(Q_{\hat{Y}|U},Q_{{U}|X})$ induced by  a randomized encoder $Q_{U|X}:\mathcal{X}\rightarrow\mathcal{P}(\mathcal{U})$ and decoder  $Q_{\hat{Y}|U}:\mathcal{U}\rightarrow\mathcal{P}(\mathcal{Y})$ is bounded by
\begin{align}
\mathcal{R}_{Y,Q_{\hat{Y}|U}}^{-1}\left( \mathcal{I}(P_X;Q_{U|X})  \right)&\leq \mathcal{R}_{Y,Q_{\hat{Y}|U}}^{-1}\left( \mathcal{I}(P_Y;Q_{U|Y})  \right) \\
&\leq  P_{\mathcal{E}}(Q_{\hat{Y}|U},Q_{{U}|X})\\
&\leq  1-\exp\left(-\mathcal{L}(Q_{\hat{Y}|U},Q_{U|X})\right),\label{eq-surrogate-bound2}
\end{align}
where $Q_{U|Y}(u|y)=\sum_{x\in\mathcal{X}}Q_{U|X}(u|x)P_{X|Y}(x|y)$ for $(u,y)\in\mathcal{U}\times\mathcal{Y}$.

\end{lemma}
\begin{IEEEproof}
The proof uses information-theoretic arguments and it is relegated to the Appendix~\ref{app:lemma-surogate}.
\end{IEEEproof}
The lower bound in the above expression states that any limitation in the mutual information between raw data and its representation will bound from below the probability of misclassification while the upper bound shows that cross-entropy loss intriduced in~\eqref{logarithmic-loss-risk} can be used as a surrogate to optimize the probability of misclassification as was also pointed out in Lemma \ref{lemma-surogate}. As a matter of fact, it appears that the probability of misclassification is  controlled by two fundamental information quantities: mutual information $\mathcal{I}(P_X;Q_{U|X})$ and the cross-entropy loss $\mathcal{L}(Q)$. As we will see later, these quantities will play a major role in the results through this paper.  

The learner's goal is  to select $Q_{U|X}$ and $Q_{\hat{Y}|U}$ by minimizing the risk~\eqref{logarithmic-loss-risk}. However, since ${P}_{XY}$ is unknown the learner cannot directly measure the risk and it is common to measure the agreement of a pair of candidates with a training data set based on the \emph{empirical risk}.

\begin{definition}[Empirical risk]
Let $\hat{P}_{XY}$ denote the empirical distribution through the training data set $\mathcal{S}_n \coloneqq \{(x_1,y_1),\dots,(x_n,y_n) \}$. The empirical risk is defined by
\begin{align}
\mathcal{L}_{\text{emp}}(Q,{\mathcal{S}_n} )&\coloneqq  \mathbb{E}_{\hat{P}_{XY}}  \left[ \ell\big(\mathbf{Q}_{U|X}(\cdot|X),\mathbf{Q}_{\hat{Y}|U}(Y | \cdot)\big) \right]\\
&=\frac1n\sum\limits_{i=1}^n \ell\big(\mathbf{Q}_{U|X}(\cdot|x_i),\mathbf{Q}_{\hat{Y}|U}(y_i | \cdot)\big).\label{eq-EM-def} 
\end{align}
\end{definition}
\begin{lemma}[Optimality of empirical decoders]\label{lemma-empirical-decoder}
Given a randomized encoder $Q_{U|X}:\mathcal{X}\rightarrow\mathcal{P}(\mathcal{U})$, define the empirical decoder w.r.t. the empirical distribution $\hat{P}_{XY}$ as:  
\begin{equation}
\hat{Q}_{Y|U}(y|u)\coloneqq \frac{\sum\limits_{x\in\mathcal{X}} Q_{U|X}(u|x)\hat{P}_{XY}(x,y)}{\sum\limits_{x\in\mathcal{X}} Q_{U|X}(u|x)\hat{P}_{X}(x)}. 
\label{eq:optimum_emp_decoder}
\end{equation}
Then, the risk can be lower  bounded uniformly over $Q_{\hat{Y}|U}:\,\mathcal{U}\rightarrow\mathcal{P}(\mathcal{Y})$ as: 
\begin{align}
\mathcal{L}_{\text{emp}}(Q_{\hat{Y}|U},Q_{U|X},{\mathcal{S}_n}) \geq   \mathcal{L}_{\text{emp}}(\hat{Q}_{Y|U},Q_{U|X},{\mathcal{S}_n}), 
\end{align}
where equality holds provided that $Q_{\hat{Y}|U}\equiv \hat{Q}_{Y|U}$, i.e., the optimal decoder is computed from the encoder and the empirical distribution as done in (\ref{eq:optimum_emp_decoder}).
\end{lemma}
\begin{IEEEproof}
The inequality follows along the lines of Lemma~\eqref{lemma-optimal-decoder} by noticing that $\mathcal{L}_{\text{emp}}(Q_{\hat{Y}|U},Q_{U|X},{\mathcal{S}_n}) =   \mathcal{D}\left(\hat{Q}_{Y|U}  \|Q_{\hat{Y}|U} |\hat{Q}_U\right)+\mathcal{L}_{\text{emp}}(\hat{Q}_{Y|U},Q_{U|X},{\mathcal{S}_n}).
$ Finally, the non-negativity of relative conditional entropy completes the proof.
\end{IEEEproof}
It is worth to mention that, due to computational and stability reasons, in many real-world applications is often preferable  to work with a restricted model class of randomized decoders (see Section \ref{subsec:restricted}) that in general may not match exactly the empirical decoder~\eqref{eq:optimum_emp_decoder}. In the light of this observation, we will consider a general class of randomized decoders. 

Since the empirical risk is evaluated on finite samples, its evaluation may be sensitive to sampling (noise) error and thus, there is the issue of generalization. It can be argued, that a key component of learning is not just the development of a representation model on the basis of finite experience, but its use to generalize to unseen experience. Clearly successful generalization necessitates the closeness (in some sense) of the selected representation and decoder models. Therefore, successful representation learning would involve successful generalization. This paper deals with the information complexity of successful generalization. The generalization gap is a measure of how an algorithm could perform on new data, i.e., data that is not available during the training phase.
\begin{definition}[Generalization gap]
Given two stochastic mappings $Q_{U|X}:\mathcal{X}\rightarrow\mathcal{P}(\mathcal{U})$ and $Q_{\hat{Y}|U}:\mathcal{U}\rightarrow\mathcal{P}(\mathcal{Y})$, the \emph{generalization gap} is defined as: 
\begin{equation}
\mathcal{E}_{\textrm{gap}}(Q,{\mathcal{S}_n}) \coloneqq\left| \mathcal{L}_{\text{emp}}(Q,{\mathcal{S}_n}) - \mathcal{L}(Q) \right|,  \label{def-gap}
\end{equation}
which represents the error incurred by the selected pair $Q=({Q}_{\hat{Y}|U},{Q}_{U|X})$ when the rule $\mathcal{L}_{\text{emp}}({Q},{\mathcal{S}_n} )$ is used to estimate the true risk $\mathcal{L}({Q})$. 
\end{definition}

\subsection{Optimizing on Restricted Classes of Randomized Encoders and Decoders}
\label{subsec:restricted}

We have already introduced the notions of representation and inference models and risk functions from which these candidates are chosen. We first introduce the Bayes risk and then the restricted classes of randomized encoders and decoders. 

\begin{definition}[Bayes risk]
The minimum cross-entropy risk over all possible encoders and decoders is called the \emph{Bayes risk} and will be denoted by $\mathcal{L}^\star $. In this case, 
\begin{align}
\mathcal{L}^\star & \coloneqq \min_{\substack{Q_{U|X}:\,\mathcal{X}\rightarrow\mathcal{P}(\mathcal{U})  \\ Q_{\hat{Y}|U}:\,\mathcal{U}\rightarrow\mathcal{P}(\mathcal{Y})}}
\mathcal{L}\big({Q}_{U|X},Q_{\hat{Y}|U}\big)  = \mathcal{H}(P_{Y|X}| P_X). \label{eq-optimal-bayes}
\end{align}
\end{definition}

\begin{definition}[Learning model]
The encoder functions are defined by $f_{\bm{\theta}}:\mathcal{X}^d \times\mathcal{Z}\longrightarrow\mathcal{U}_{\bm{\theta}}^m$, where $\mathcal{X}$ is the finite input alphabet with cardinality $| \mathcal{X}|$, $d$ is a positive integer number, $\bm{\theta}\in\Theta\subset \mathbb{R}^{d_{\Theta}}$ denotes unknown  parameters to be optimized, $Z$ is a random variable taking values on a finite alphabet $\mathcal{Z}$ with probability $P_Z$ whose role is to randomize encoders (decoders) and $\mathcal{U}_{\bm{\theta}}\subset [0,1]$ is the alphabet corresponding to the hidden representation  which satisfies $| \mathcal{U}_{\bm{\theta}}|\leq | \mathcal{X}|^d\cdot | \mathcal{Z}|$. For notation convenience,  we let $\mathcal{X} \equiv \mathcal{X}^d$ and  $\mathcal{U}\equiv\mathcal{U}_{\bm{\theta}}^m$ and denote this class as:
$$
\mathcal{F}_E \coloneqq \left\{ Q_{U|X}(u|x) = \mathbb{E}_{P_Z} \big[\mathds{1} [u=f_{\bm{\theta}}(x,Z)] \big] \ : \   \bm{\theta}\in\Theta\right\}.
$$
The decoder functions can be defined as $g_{\bm{\gamma}}:\mathcal{U}_{\bm{\theta}}^m \times\mathcal{Z}\longrightarrow\mathcal{Y}$ where  $\bm{\gamma}\in\Gamma\subset \mathbb{R}^{d_D}$ denotes the unknown parameters. Similarly, we let 
$$
\mathcal{F}_D \coloneqq \left\{ Q_{\hat{Y}|U} (y|u) = \mathbb{E}_{P_Z} \big[\mathds{1} [y=g_{\bm{\gamma}}(u,Z)] \big] \ : \   \bm{\gamma}\in\Gamma\right\}.
$$
It is clear that for every $(\bm{\theta},\bm{\gamma})$, $\bm{\theta}\mapsto Q_{U|X}\in \mathcal{F}_E$ induces a randomized encoder while $\bm{\gamma}\mapsto Q_{\hat{Y}|U} \in \mathcal{F}_D$ induces a randomized decoder.
\end{definition}

In order to simplify subsequent  analysis we will assume the following conditions over the possible data pmf and over the families $\mathcal{F}_E$ and $\mathcal{F}_D$ of encoders and decoders:
\begin{assumptions}[Restricted model class]\label{asumption}
 We assume that alphabets $\mathcal{X}$, $\mathcal{Y}$ are of arbitrary large size but finite. Furthermore, there exists  $\eta>0$ such that:
\begin{itemize}
\item For every $Q_{\hat{Y}|U} \in \mathcal{F}_D $, $Q_{\hat{Y}|U}(y_{\min}|u) \coloneqq \min_{y\in\mathcal{Y}}  Q_{\hat{Y}|U}(y|u)\geq \eta$ for each $u\in\mathcal{U}$;
\item The unknown $P_{XY}$ satisfies $P_{X}(x_{\min}) \coloneqq \min_{x\in\mathcal{X}} P_X(x)\geq \eta$ and $P_{Y}(y_{\min}) \coloneqq \min_{y\in\mathcal{Y}} P_Y(y)\geq \eta$.
\end{itemize}
\end{assumptions}

\begin{definition}[Empirical risk minimization] 
The methodology of empirical risk minimization is one of the most straight-forward, yet it is usually efficient provided that the chosen model class $\mathcal{F}\coloneqq  (\mathcal{F}_E,\mathcal{F}_D)$ is sufficient restrictive~\cite{vapnik98statlearn}. The learner chooses a pair $(\hat{Q}^\star_{U|X},\hat{Q}^\star_{\hat{Y}|U})\in \mathcal{F} $ that minimizes the empirical risk:
\begin{equation}
\mathcal{L}_{\text{emp}}\big(\hat{Q}^\star_{U|X},\hat{Q}^\star_{\hat{Y}|U},{\mathcal{S}_n} \big) \leq    \mathcal{L}_{\text{emp}}(\hat{Q}_{U|X},\hat{Q}_{\hat{Y}|U},{\mathcal{S}_n}),\label{eq-empirical-risk}
\end{equation}
for all $(\hat{Q}_{U|X},\hat{Q}_{\hat{Y}|U})\in \mathcal{F}$.
\end{definition}
Moreover, it is possible to minimize a surrogate of the true risk: 
\begin{equation}\label{eq:tradeoff-empgen}
\mathcal{L}({Q})\leq \mathcal{L}_{\text{emp}}(Q,{\mathcal{S}_n})+\mathcal{E}_{\textrm{gap}}(Q,{\mathcal{S}_n}),\,\,\,
\end{equation}
which depends on the empirical risk and the so-called generalization gap, respectively. Expression~\eqref{eq:tradeoff-empgen} states that an adequate selection of the encoder-decoder pair should be chosen in order to minimize the empirical risk and the generalization gap simultaneously.  It is reasonable to expect that the assumption that the optimal encoder-decoders pairs achieving the minimal risk in~\eqref{eq-optimal-bayes} do not belong to our restricted class of models $\mathcal{F}$, so the learner may want to enlarge the model classes $\mathcal{F}$ as much as possible. However, this could induce a larger value of the generalization gap, which could lead to a tradeoff between these two fundamental quantities.

\subsection{Main Focus of this Work}
We study informational bounds on the generalization gap~\eqref{def-gap}. More precisely, the goal is to find the learning rate: $\epsilon_{n}({Q},\mathcal{S}_n, \gamma_n)$ such that
\begin{equation}
\mathbb{P}\left( \mathcal{E}_{\textrm{gap}}(Q,{\mathcal{S}_n}) > \epsilon_n({Q},\mathcal{S}_n, \gamma_n)\right)\leq \gamma_n,  
\end{equation}
for every given pair ${Q}=({Q}_{\hat{Y}|U},{Q}_{U|X})\in\mathcal{F}$ and  some $\gamma_n \rightarrow 0$ as $n\rightarrow \infty$. We will further study PAC style bounds that allow us to compute with precision the rate $\epsilon_{n}({Q},\mathcal{S}_n, \gamma_n)$ that satisfies: 
\begin{equation}
\mathbb{P}\left( \mathcal{L}({Q})\leq \mathcal{L}_{\text{emp}}\big({Q},{\mathcal{S}_n} \big)+ \epsilon_n({Q},\mathcal{S}_n, \gamma_n)\right)\geq  1-\gamma_n.   
\end{equation}
We will further introduce the notion of information complexity and study properties and implications for practical algorithms minimizing the corresponding regularized risk:
\begin{equation}
\mathcal{L}_{\text{emp}}\big({Q},{\mathcal{S}_n} \big) + \lambda\cdot\epsilon_n({Q},\mathcal{S}_n, \gamma_n),
\label{eq:empirical_with_reg}
 \end{equation}
for some adequate multiplier $\lambda\geq 0$. With the above background in mind, we can now start to develop our main machinery, as well as its consequences in representation learning theory.

\section{Informational Bounds on the Generalization Gap}\label{Section-3}

In this section, we first present our main result in Theorem~\ref{thm:regularizar} which is a bound on the generalization gap~\eqref{def-gap} with probability at least $1-\delta$, as a function of every randomized encoder and decoder pair $(Q_{U|X},Q_{\hat{Y}|U})$ and probability  distribution $P_{XY}$. In particular, we  will show that the mutual information between raw data and its representation control the learning rate with an order $\mathcal{O}\left(\frac{\log(n)}{\sqrt{n}}\right)$, which leads to an informational PAC style generalization error bound. From this perspective, we motivate the notion of information complexity and discuss implications for model selection, variational autoencoders and information bottleneck method.  Furthermore, in Corollary~\ref{corollary2} we show that the classification error is upper and lower bounded as functionals of the information complexity. 

\subsection{Informational Bound on the Generalization Gap}

\begin{theorem}[Informational bound]\label{thm:regularizar}
Let $\mathcal{F} \coloneqq (\mathcal{F}_E,\mathcal{F}_D) $ be a class of decoders and decoders. Then, for every  $P_{XY}$ and every $\delta\in(0,1)$, with probability at least $1-\delta$ over the choice of $\mathcal{S}_n\sim P_{XY}^n$ the following inequality holds $\forall\, Q=(Q_{U|X},Q_{\hat{Y}|U})\in\mathcal{F}$:
\begin{align}
&\mathcal{E}_{\textrm{gap}}(Q,\mathcal{S}_n)\leq A_\delta\sqrt{\mathcal{I}(\hat{P}_X;Q_{U|X})}\cdot\frac{\log(n)}{\sqrt{n}}+\frac{C_\delta}{\sqrt{n}} + \frac{B_\delta{\Lambda}\big(Q_{U|X},\hat{Q}_{Y|U},Q_{\hat{Y}|U}\big)} {\sqrt{n}}+\mathcal{O}\left(\frac{\log(n)}{n}\right),\label{eq-main-bound-GG}
\end{align} 
where we defined the decoder efficiency:
\begin{align}
{\Lambda}&\big(Q_{U|X},\hat{Q}_{Y|U},Q_{\hat{Y}|U}\big) \coloneqq  \sqrt{\frac{1}{|\mathcal{X}||\mathcal{Y}|} \sum\limits_{\forall (x,y)} \sum\limits_{\forall (x^\prime,y^\prime)} \big[\Delta(x,y) - \Delta(x^\prime,y^\prime)  \big]^2 }
\end{align}
with $\Delta(x,y) \coloneqq \ell\big(\mathbf{Q}_{U|X}(\cdot|x),\mathbf{Q}_{\hat{Y}|U}(y | \cdot)\big) -\ell\big(\mathbf{Q}_{U|X}(\cdot|x),\mathbf{\hat{Q}}_{Y|U}(y | \cdot)\big)$, and where $(A_\delta,B_\delta,C_\delta)$ are  universal constants given by:
\begin{IEEEeqnarray}{rCl}
A_\delta& \coloneqq&  \frac{\sqrt{2}B_\delta}{P_{X}(x_{\min})}\left(1+\sqrt{\frac{1}{|\mathcal{X}|}}\right),\\
B_\delta&\coloneqq& \left(1+\sqrt{\log\left(\frac{|\mathcal{Y}|+4}{\delta}\right)}\right),\\
C_\delta& \coloneqq & 2|\mathcal{U}|e^{-1}+B_\delta\sqrt{|\mathcal{Y}|} \log \left(\frac{|\mathcal{U}|}{P_{Y}(y_{\min})}\right).
\end{IEEEeqnarray}
\end{theorem}
\begin{IEEEproof}
The proof is relegated to Section~\ref{sec:proof_main_theorem}.	
\end{IEEEproof}

The importance of this bound is that the left hand side is a quantity that involves the empirical mutual information between raw data $X$ and its randomized representation $U(X)$ that can be understood as a ``measure of information complexity'' of the representations, and  the right hand side term could be seen as a measure of the decoder efficiency, both terms with different scaling rates $n^{-1/2}\log(n)$ and $n^{-1/2}$. It should be pointed out that since any randomized decoder $Q_{\hat{Y}|U}$ other than the empirical decoder $\hat{Q}_{Y|U}$ is suboptimal (see Lemma~\ref{lemma-empirical-decoder}), ${\Lambda}\big(Q_{U|X},\hat{Q}_{Y|U},Q_{\hat{Y}|U}\big)$ is a kind of similarity measure between the losses associated to each of these decoders, which is vanishes by selecting $\hat{Q}_{Y|U}\approx Q_{\hat{Y}|U}$. Therefore the inequality in~\eqref{eq-main-bound-GG} is a result that can be applied to an arbitrary pair of randomized encoder and decoder. The remaining issue is merely how to interpret this information-theoretic bound.

\subsection{Information Complexity of Representations}\label{sec-info-complex}

We could think of the most significative term in the upper bound~\eqref{eq-main-bound-GG} as an information complexity cost of data representations, which depends only on the data samples and on the selected randomized encoder from the restricted model. Suppose we are given with a set of different model classes for the randomized encoders: 
\begin{align*}
\mathcal{F}_E^{(k)}\coloneqq&\left\{ Q_{U|X}(u|x) = \mathbb{E}_{P_Z} \big[\mathds{1} [u=f_{\bm{\theta}}(x,Z)] \big] \,:\,   \bm{\theta}=(\theta_1,\dots,\theta_k) \in\Theta_k,\,\, P_Z\in\mathcal{P}_k(\mathcal{Z}) \right\},
\end{align*}
$k=[1:K]$, where there are two kinds of parameters: a structure parameter $k$ and real-value parameters $\theta$, whose parameters depend on the structure, e.g., $\Theta_k$ may account for different number of layers or non-linearities while $\mathcal{P}_k(\mathcal{Z})$ indicates different kind of noise distributions. Our result in Theorem~\ref{thm:regularizar} motivates the following model selection principle for learning compact representations: 
\begin{quote}
{\it
Find a structure parameter $k$ and real-value parameters $\bm{\theta}$ for the observed data $\mathcal{S}_n$ with which the corresponding data representation can be encoded with the shortest code length: 
\begin{align}
\inf\limits_{\bm{\theta}\in\Theta_k\,, \,k =[1:K]}\,&\left[ \mathcal{L}_{\text{emp}}\left(Q_{U|X}^{(\bm{\theta},k)},Q_{\hat{Y}|U}^{(\bm{\theta},k)},{\mathcal{S}_n}\right) +\lambda \cdot \sqrt{\mathcal{I}\left(\hat{P}_X;Q_{U|X}^{(\bm{\theta},k)}\right)}  \right],\label{eq-principle-IT}
\end{align}
where the mutual information penalty term indicates the minimum of the expected redundancy between the minimum code-length\footnote{As is well known in information theory, the shortest expected code length achievable by a uniquely decodable code under a known data source~\cite{cover}.} (measured in bits) $-\log Q_{U|X}^{(\bm{\theta},k)}(\cdot|x)$ to encode the data representations under a known data source and the best code-length $-\log Q_{U}(\cdot)$ chosen to encode the data representations without knowing the input samples: 
}
\end{quote} 
\begin{align}
\mathcal{I}\left(\hat{P}_X;Q_{U|X}^{(\bm{\theta},k)}\right)=& \min_{Q_U \in\mathcal{P}(\mathcal{U})}\mathbb{E}_{\hat{P}_X}  \mathbb{E}_{Q_{U|X}^{(\bm{\theta},k)}}\left[- \log Q_{U}(U)+ \log  Q_{U|X}^{(\bm{\theta},k)}(U|X)  \right].\label{eq-complexity-data-repre}
\end{align}
This informational principle combines the empirical cross-entropy risk~\eqref{eq-EM-def} with the ``information complexity'' of the selected  encoder~\eqref{eq-complexity-data-repre} as being a regularization that acts as a sample-dependent penalty against \emph{overfitting}.  One may view \eqref{eq-complexity-data-repre} as a possible means of comparing the appropriateness of distinct models (e.g., number of layers and amount of noise), after a parametric choice has been selected.

The coding interpretation of the penalty term in \eqref{eq-principle-IT} is that the length of the description of the representations themselves can be quantified in the same units as the code length in data compression, namely, bits. In other words, for each data sample ${x}$, a randomized encoder can induce different types of representations $U(x)$ with expected information length given by $\mathcal{H}\left(Q_{U|X}(\cdot|x)\right)$. When this representation has to be encoded without knowing $Q_{U|X}$ since $x$ is not given to us (e.g. a communication problem where the sender wishes to communicate the representations only), the required average length of an encoding distribution $Q_{U}$ results in $\mathbb{E}_{Q_{U|X}} \big[-\log Q_{U}(U) \big]$. In this sense, expression~\eqref{eq-principle-IT} suggests to select encoders that allow us to then encode representations efficiently.  Interestedly, this  is closely related to the celebrated minimum description length (MDL) method for density estimation~\cite{Rissanen:1978:PMS:2233043.2233262,Grnwald:2005:AMD:1051706}. However, two fundamental differences  between these principles is that our information complexity~\eqref{eq-complexity-data-repre} follows from the generalization gap and measures the amount of information convey by the representations relative to an encoder model, as opposed to the model parameters of the encoder itself.

The information-theoretic significance of \eqref{eq-complexity-data-repre} goes beyond  a simply regularization term since it leads to us to introduce the fundamental notion of \emph{encoder capacity}. This key idea of encoder capacity is made possible by Theorem~\ref{thm:regularizar}  that connects mathematically the generalization gap to the information complexity which is intimately related to the number of distinguishable samples from representations. Notice that the information  complexity can be upper bounded with \emph{Jensen inequality} as: 
\begin{equation}
\mathcal{I}\left(\hat{P}_X;Q_{U|X}\right)\leq\frac{1}{n^2}\sum_{i=1}^{n}\sum_{j=1}^n \mathcal{D}\left(Q_{U|X}(\cdot|x_i) \big\| Q_{U|X}(\cdot|x_j)\right),
\end{equation}
where $\left\{x_i\right\}_{i=1}^n$ are the training examples from the dataset $\mathcal{S}_n$. This bound is measuring the average degree of closeness between the corresponding representations for the different sample inputs. When two distributions, $Q_{U|X}(\cdot|x_i)$ and $Q_{U|X}(\cdot|x_j)$ are very close to each other, i.e.,  $Q_{U|X}$  assigns high likelihood to similar representations corresponding to different inputs $x_i\neq x_j$, they do not contribute so much to the complexity of the overall representations. In other words, the more sample inputs an encoder can differentiate, the more patterns it can fit well so the larger the mutual information will be and thus the risk of overfitting. This observation suggests the complexity of a representation model with respect to a sample dataset can be measured in terms of the number of data samples  that essentially yields different (distinguishable) representations. Inspired by the concept of \emph{stochastic complexity}~\cite{Grnwald:2005:AMD:1051706}, we introduce below the notion of  \emph{encoder capacity} to measure the complexity of a representation model:
\begin{definition}[Capacity of randomized encoders]\label{definition-encoder-capacity}
The \emph{encoder capacity} $\mathcal{C}_e$ of a randomized encoder $Q_{U|X}$ with respect to a sample set $\mathcal{A}\subseteq \mathcal{X}$ is defined as: 
\begin{IEEEeqnarray}{rCl}
\mathcal{C}_e(\mathcal{A},Q_{U|X}) &\coloneqq & \max\limits_{\psi\,:\,\mathcal{U}\rightarrow \mathcal{A}} \log \left(\sum\limits_{u\in\mathcal{U}} Q_{U|X}\big(u|\psi(u)\big) \right) = \log \left| \mathcal{A} \right| - \log  \left(\frac{1}{1 -\varepsilon }\right),\label{eq-capacity-encoder-expression}
\end{IEEEeqnarray}
\begin{IEEEeqnarray}{rCl}
\varepsilon &\coloneqq& 
\min\limits_{\psi\,:\,\mathcal{U}\rightarrow \mathcal{A}}  \frac{1}{| \mathcal{A} |} \sum_{x\in \mathcal{A}}  \sum_{u\in\mathcal{U}}{Q}_{U|X}
\left(u|x\right)\mathds{1}\left[ \psi(u)\neq x \right] \leq 1-\frac{1}{|\mathcal{A}|}.  \label{eq-capacity-encoder-expression2}
\end{IEEEeqnarray}
\end{definition}

The argument of the logarithm in the second term of~\eqref{eq-capacity-encoder-expression} represents the probability to  distinguish samples from their representations $1-\varepsilon$, i.e., the average probability that estimated samples via the maximum-likelihood estimator $\psi(\cdot)$ from $Q_{U|X}$ be equal to the true samples. Therefore,  the encoder capacity is the logarithm of the number of total samples minus a term that depends on the misclassification probability of the input samples from their representations. When $\varepsilon$ is small, then $\mathcal{C}_e(\mathcal{A},Q_{U|X}) \approx \log | \mathcal{A} | -\varepsilon$  and thus, all samples are perfectly distinguishable. The following proposition gives simple bounds\footnote{Notice that it is possible  to provide better bounds on $\varepsilon$ by relaying on the results in~\cite{1055007}.  However, we preferred simplicity to ``tightness'' since the purpose  of Proposition~\ref{lemma-capacity-encoder} is to link the encoder capacity and the information complexity.} on the encoder capacity from the information complexity~\eqref{eq-complexity-data-repre}, which relates to the generalization gap: 

\begin{proposition} \label{lemma-capacity-encoder}
Let $Q_{U|X}$ be an encoder distribution and $\hat{P}_X$ be an empirical distribution with support $\mathcal{A}_n \equiv \textrm{supp}(\hat{P}_X)$. Then, the information complexity and the encoder capacity satisfy: 
\begin{IEEEeqnarray}{rCl}
\mathcal{C}_e \big(Q_{U|X},\mathcal{A}_n\big)&=&\log \left| \mathcal{A}_n \right| -  \log  \left(\frac{1}{1 -\varepsilon }\right)\label{propo-capacity1}\\
g^{-1}\left(\log \left| \mathcal{A}_n \right| - \mathcal{I}\big(\hat{P}_X;Q_{U|X} \big) \right) &\leq& \varepsilon \leq \frac12 \left( \log \left| \mathcal{A}_n \right| - \mathcal{I}\big(\hat{P}_X;Q_{U|X} \big) \right),  \label{propo-capacity2}
\end{IEEEeqnarray} 
where $\varepsilon$ is defined by~\eqref{eq-capacity-encoder-expression2} with respect to $\mathcal{A}_n$ and, for $0\leq t\leq 1$,  
\begin{equation}
g(t) \coloneqq t \cdot\log \left(\left| \mathcal{A}_n \right| -1\right) + h(t)
\end{equation}
with $h(t) \coloneqq -t\log(t)-(1-t)\log(1-t)$ and $0\log 0 \coloneqq 0$. Furthermore, 
\begin{equation}
\mathcal{I}\big(\hat{P}_X;Q_{U|X} \big) \leq \mathcal{C}_e\big(Q_{U|X},\mathcal{A}_n\big).  \label{eq-upper-bound-IC}
\end{equation}
\end{proposition}
\begin{IEEEproof}
The proof is relegated to  Appendix~\ref{app:lemma-capacity-encoder}.
\end{IEEEproof}

Despite the encoder capacity~\eqref{eq-capacity-encoder-expression} is an important  theoretical concept that is very appealing because it connects generalization with the probability of distinguishing inputs from their representations, the derivation of analytic expressions for relevant classes of randomized encoders can be very challenging in practice. In this sense, inequality~\eqref{propo-capacity2} allows us to bound this capacity via the information complexity, as will be done in Section~\ref{section-dropout-complex}.
\begin{remark}
	In Proposition~\ref{lemma-capacity-encoder}, the function  $g^{-1}(t)\coloneqq 0$ for $t<0$ and, for $0<t< \log \left| \mathcal{A}_n \right|$, $g^{-1}(t)$ is a solution of the equation $g(\varepsilon)=t$ with respect to $\varepsilon\in\big[0,  1-1/\left| \mathcal{A}_n \right| \big]$; this solution exists since the function $g$ is continuous and increasing on  $\big[0, 1-1/\left| \mathcal{A}_n \right|\big]$ and $g(0)=0$, $g\big(  1-1/\left| \mathcal{A}_n \right| \big) =  \log \left| \mathcal{A}_n \right|$. 
\end{remark}
\begin{remark}[Generalization requires learning invariant representations]\\
An important consequence of  the lower bound in~\eqref{propo-capacity2} in Proposition~\ref{lemma-capacity-encoder}  is that by limiting the information complexity, i.e., by controlling the generalization gap according to the criterion~\eqref{eq-principle-IT}, we bound from below the error probability of distinguishing input samples from their representations. In other words, from expression~\eqref{propo-capacity1} and  Theorem~\ref{thm:regularizar} we can conclude that encoders inducing a large misclassification probability on input samples from their  representations, i.e., different inputs must share similar representations, are expected to achieve better generalization. Specifically, this also implies formally that we only need to enforce invariant representations to control the encoder capacity (e.g., injecting noise during training), from which the generalization is upper bounded naturally thanks to Theorem~\ref{thm:regularizar} and the connection with the information complexity. However, there is a sensitive tradeoff between the amount of noise (enforcing both invariance and generalization) and the minimization of the cross-entropy loss, as shown in Section~\ref{section-dropout-complex}. Additionally, it is not difficult to show from data-processing inequality that stacking noisy encoder layers reenforce increasingly invariant representations since distinguishing inputs from their representations  becomes harder  --or equivalently the encoder capacity decreases-- as the network is deeper. Interestedly, our  conclusions here are consistent with previous works in~\cite{6522407, Achille_Soatto_2017}.
\end{remark}


\subsection{A Note on Computation of Information Complexity}

In the reminder, we  concentrate in providing bounds on the information complexity~\eqref{eq-complexity-data-repre}.  

\emph{Relationship between information complexity and metric entropy.} An alternative formulation of  the information complexity follows by using the metric entropy of the model class. Consider the interesting identity derived in~\cite[Theorem 1]{yang1999}:
\begin{align}
\mathcal{I}&(\hat{P}_X;Q_{U|X})\leq \min\limits_{\mathcal{G}\subseteq \textrm{supp}(\hat{P}_X)} \left\{\log |\mathcal{G} | +\max_{x\in\{x_1\cdots x_n \}} \min_{x^\prime\in \mathcal{G}} \mathcal{D}\big( Q_{U|X}(\cdot|x) \| Q_{U|X}(\cdot|x^\prime) \big)\right\},\label{eq-covering}
\end{align}
where $\{Q_{U|X}(\cdot|x)\}_{x\in\mathcal{G}}$ can be viewed as an approximation of the set $\{Q_{U|X}(\cdot|x_i)\}_{i=\{1,\dots,n \}}$.  Let $\mathcal{G}_{\epsilon}$ be the set of minimum cardinality  satisfying: $\mathcal{D}\big( Q_{U|X}(\cdot|x) \| Q_{U|X}(\cdot|x^\prime) \big) \leq \epsilon$ for each ${x\in\{x_1,\dots,x_n \}}$ and some ${x^\prime\in \mathcal{G}_{\epsilon}}$.  Then,  $\{Q_{U|X}(\cdot|x)\}_{x\in\mathcal{G}_{\epsilon}}$ is an $\epsilon$-cover of the set $\{Q_{U|X}(\cdot|x)\}_{x\in\{x_1\cdots x_n \}}$ with metric entropy  $\log |\mathcal{G}_{\epsilon}|$. The upper bound \eqref{eq-covering} measures the spread of the randomized encoder $\{Q_{U|X}(\cdot|x)\}_{x\in\mathcal{X}}$. When the probabilities $\{Q_{U|X}(\cdot|x_i)\}_{i=\{1,\dots,n \}}$ are tighly packed it is possible to approximate them well with a smaller set and the bound in (\ref{eq-covering}) will be small.



\emph{Deep randomized encoders}. In order to further simplify expression~\eqref{eq-complexity-data-repre} in presence of multiple layers, we can focus on simple methods for bounding the total information complexity. Let $Q_{U|X}$ be a ``deep randomized encoder'' that is $Q_{U|X}$ is defined as the marginal distribution over a stack of randomized encoders $Q_{U_1|X}:\mathcal{X}\rightarrow\mathcal{P}(\mathcal{U}_1), \dots, Q_{U_K|U_{K-1}}:\mathcal{U}_{K-1}\rightarrow\mathcal{P}(\mathcal{U}_K)$ with $Q_{U|X}=Q_{U_K|X}$, where
\begin{equation}
Q_{U_1,\cdots,U_K|X}=Q_{U_1|X}Q_{U_2|U_1}\cdots Q_{U_K|U_{K-1}}
\end{equation}
From data processing inequality~\cite{cover}, it follows that $\mathcal{I}(\hat{P}_X;Q_{U_1|X}) \leq  \mathcal{I} \big(Q_{U_{1}};Q_{U_2|U_{1}}\big) \leq \cdots \leq  \mathcal{I} \big(Q_{U_{K-1}};Q_{U_K|U_{K-1}}\big)$ and thus, for all $\lambda_k\in [0,1]$ with $\sum_{k=1}^K \lambda_k=1$, we have: 
 \begin{equation}
\mathcal{I}(\hat{P}_X;Q_{U|X}) \leq  \sum\limits_{k=1}^K \lambda_k\cdot\mathcal{I}\big(Q_{U_{k-1}};Q_{U_k|U_{k-1}}\big), \label{eq-bound-multilayer}
\end{equation}
where $Q_{U_{0}} \equiv  \hat{P}_X$. Expression~\eqref{eq-bound-multilayer} provides an upper bound to the overall  information complexity as a function of per-layer complexities.

\subsection{Connections to Related  Algorithms}

In this subsection we study how the objective given by expression~\eqref{eq-principle-IT}, which was motivated from our bound on the generalization gap, is indeed similar to well-known regularized risks available in the literature. 

It is not hard to see that if the set of all possible (universal) randomized decoders is allowed, then the optimal decoder, than minimize the upper bound~\eqref{eq:tradeoff-empgen}, is given by ${Q}_{\hat{Y}|U} \equiv \hat{Q}_{Y|U}$,  indicating that the generalization capabilities depend mostly on the representation model\footnote{There are many practical learning classes in which the decoder is given and only the randomized encoder $Q_{U|X}$ has to be determined  (e.g., \emph{softmax regression}).}. The class of universal decoders leads to the following PAC style generalization error bound. 

\begin{corollary}[Universal class of decoders]\label{lem:bottleneck}
Let $\mathcal{F} \coloneqq (\mathcal{F}_E, \mathcal{F}_D) $ and  $\mathcal{F}_D\coloneqq \{Q_{\hat{Y}|U}:\mathcal{U}\rightarrow\mathcal{P}(\mathcal{Y})  \}$ be the class of universal decoders. Then, for every $P_{XY}$ and every $\delta\in(0,1)$, with probability at least $1-\delta$ over the choice of $\mathcal{S}_n\sim P_{XY}^n$ the following inequality holds: 
\begin{align}
&\hspace{-0.4cm}\mathcal{L}(\hat{Q}_{Y|U},Q_{U|X})\leq  \mathcal{H}(\hat{Q}_{Y|U}|\hat{Q}_{U}) +A_\delta\sqrt{\mathcal{I}(\hat{P}_X;Q_{U|X})}\cdot \frac{\log(n)}{\sqrt{n}}+\frac{C_\delta}{\sqrt{n}}+\mathcal{O}\left(\frac{\log(n)}{n}\right).
\label{eq:loss_only_encoder}
\end{align}
\end{corollary}
An interesting connection between the empirical risk minimization of the cross-entropy  loss and the Information Bottleneck (IB) principle~\cite{Tishby1999information,2016arXiv161101353A,DBLP:journals/corr/AlemiFD016} arises which motivates formally  the following algorithm. 

\begin{definition}[Information Bottleneck algorithm]\label{def-IB-algo}
A learning algorithm inspired by the IB method~\cite{Tishby1999information} consists in finding $Q_{U|X}\in \mathcal{F}_E$ that minimizes over the training set choice $\mathcal{S}_n\sim P_{XY}^n$ the functional: 
\begin{equation}
\hspace{-0.2cm}\mathcal{L}_{\textrm{IB}}^{(\lambda)}(Q_{U|X}, \mathcal{S}_n) \coloneqq  \mathcal{H}(\hat{Q}_{Y|U}|\hat{Q}_U)+\lambda\cdot \mathcal{I}(\hat{P}_X;{Q}_{U|X}) ,  \label{eq-IB-function}
\end{equation}
for a suitable multiplier $\lambda>0$, where
\begin{align}
\hat{Q}_U(u)&\coloneqq \sum\limits_{x\in\mathcal{X}} Q_{U|X}(u|x)\hat{P}_X(x),\\ \hat{Q}_{Y|U}(y|u)&\coloneqq\frac{\sum\limits_{x\in\mathcal{X}} Q_{U|X}(u|x)\hat{P}_{XY}(x,y)}{\hat{Q}_U(u)}. \label{eq-empirical-decoder}
\end{align}
\end{definition}
This algorithm optimizes a tradeoff between $ \mathcal{H}(Q_{Y|U}|\hat{Q}_U)$ and an information-based regularization term $\mathcal{I}(\hat{P}_X;Q_{U|X})$. Interestingly, the resulting regularized empirical risk suggested by~\eqref{eq:loss_only_encoder} can be seen as an optimization of the IB method from the empirical distribution~\eqref{eq-IB-function} but based on the square-root of the mutual information in expression~\eqref{eq-IB-function}. 

Additionally, we observe that by selecting an arbitrary $\tilde{Q}_U\in\mathcal{P}(\mathcal{U})$ in~\eqref{eq-complexity-data-repre}\footnote{The equality is achieved when $\tilde{Q}_U(u)=\hat{Q}_U(u)=\sum_{x\in\mathcal{X}}\hat{P}_X(x)Q_{U|X}(u|x)$ for all $u\in\mathcal{U}$.}, the next inequality holds:
\begin{align}
\mathcal{L}(\hat{Q}_{Y|U},Q_{U|X})&\leq\mathcal{H}(\hat{Q}_{Y|U}|\hat{Q}_U)+\lambda\cdot \sqrt{ \mathcal{D}\big(Q_{U|X} \| \tilde{Q}_U|\hat{P}_{X} \big)}\nonumber\\
&\equiv   \mathcal{L}^{(\lambda)}_{\textrm{VA}}(Q_{U|X},\tilde{Q}_U,\mathcal{S}_n). \label{eq-expression-VA}
\end{align}
The new surrogate function~\eqref{eq-expression-VA}, denoted by $ \mathcal{L}^{(\lambda)}_{\textrm{VA}}(Q_{U|X},\mathcal{S}_n)$, shares a lot of in common with a slightly more general form of variational autoencoders (VAEs)~\cite{kingma2013auto} and its variation of the recently introduced Information Dropout (ID)~\cite{2016arXiv161101353A,DBLP:journals/corr/AlemiFD016}, where the latent space is regularized using a prior $\tilde{Q}_U$. Therefore, the mathematical connection stated by Theorem~\ref{thm:regularizar} shows that the algorithm in Definition~\ref{def-IB-algo} as well as VAEs and ID are slightly different but related ways to control the generalization gap. 

\subsection{The Interplay Between the Misclassification Probability and the Information Complexity}

Theorem~\ref{thm:regularizar} provides an upper bound (with probability $1-\delta$) on the true risk $\mathcal{L}$ and in Lemma~\ref{lemma-surogate-representations} we have related it to the ultimate quantity of interest that is the probability of misclassification $P_{\mathcal{E}}$ of our classifier. Indeed, this bound is a function of $(Q_{\hat{Y}|U},Q_{{U}|X})$ and the training dataset $\mathcal{S}_n$ that, according to (\ref{eq:empirical_with_reg}), suggests to use the information complexity as a regularizer.  Here we explore the impact of the involved information quantities in the misclassification probability. Our result relies strongly on  Lemma~\ref{lemma-surogate-representations} from which we can obtain a lower bound on $P_{\mathcal{E}}$ (with probability $1-\delta$) that is also a function of $(Q_{\hat{Y}|U},Q_{{U}|X})$ and the training dataset $\mathcal{S}_n$. This is summarized in the proposition below:

\begin{proposition}\label{lemma:lower_emp_pe}
Let $\mathcal{F} \coloneqq (\mathcal{F}_E,\mathcal{F}_D) $ be a model class of randomized encoders and decoders. Then, for every  $P_{XY}$ and every $\delta\in(0,1)$ with probability at least $1-\delta$ over the choice of $\mathcal{S}_n\sim P_{XY}^n$, the following inequality holds $\forall\, (Q_{U|X},Q_{\hat{Y}|U})\in\mathcal{F}$:
\begin{align}
&P_{\mathcal{E}}(Q_{\hat{Y}|U},Q_{{U}|X})\geq \mathcal{R}_{Y,Q_{\hat{Y}|U}}^{-1}\left( \mathcal{I}(\hat{P}_Y;\hat{Q}_{U|Y})  \right)+\left(\frac{E_\delta}{\sqrt{n}}\right)\frac{\partial \mathcal{R}_{Y,Q_{\hat{Y}|U}}^{-1}(I)}{\partial I}\Big |_{I=\mathcal{I}(\hat{P}_Y;\hat{Q}_{U|Y})}\hspace{-0.1cm}+o\left(\frac{1}{\sqrt{n}}\right),\label{eq-lower-error-classifier}
\end{align}
where $\hat{Q}_{U|Y}(u|y)=\sum_{x\in\mathcal{X}}Q_{U|X}(u|x)\hat{P}_{X|Y}(x|y)$ for $(u,y)\in\mathcal{U}\times\mathcal{Y}$ and 
\begin{equation*}
E_\delta \coloneqq  \left(1+\sqrt{\log\left(\frac{1}{\delta}\right)}\right)\sqrt{|\mathcal{X}||\mathcal{Y}|}\left[\log\left(\frac{1}{P_Y(y_{\min})}\right)-1\right].
\end{equation*}
\end{proposition}
\begin{IEEEproof}
The proof is relegated to  Appendix~\ref{app:lemma-lower_bound_pe}.
\end{IEEEproof}

\begin{remark}
Notice that the lower bound in~\eqref{eq-lower-error-classifier} is decreasing in the argument of $\mathcal{R}_{Y,Q_{\hat{Y}|U}}^{-1}(\cdot)$ which jointly with $\mathcal{I}(\hat{P}_Y;\hat{Q}_{U|Y})\leq  \mathcal{I}(\hat{P}_X; {Q}_{U|X})$, and using Theorem~\ref{thm:regularizar} and Proposition~\ref{lemma:lower_emp_pe}, allow us to have that $\forall Q=(Q_{U|X},Q_{\hat{Y}|U})\in\mathcal{F}$: 
\begin{IEEEeqnarray*}{rCl}
P_{\mathcal{E}}(Q)&\geq & \mathcal{R}_{Y,Q_{\hat{Y}|U}}^{-1}\left( \mathcal{I}(\hat{P}_X; {Q}_{U|X})  \right)+ \mathcal{O}\left(\frac{1}{\sqrt{n}}\right),\\
\mathcal{E}_{\textrm{gap}}(Q,\mathcal{S}_n) &\leq & A_\delta\sqrt{\mathcal{I}(\hat{P}_X;Q_{U|X})}\cdot\frac{\log(n)}{\sqrt{n}} +  \mathcal{O}\left(\frac{1}{\sqrt{n}}\right), 
\end{IEEEeqnarray*}
with probability at least $1-\delta$ over the choice of $\mathcal{S}_n\sim P_{XY}^n$.  In other words, by restricting the information complexity we can control the learning rate but at a cost of limiting the performance of our classifier. This exposes the existence of an interesting tension between the classification error and the generalization gap via the information complexity. 
\end{remark}

Additionally, by using Theorem~\ref{thm:regularizar} together with  the upper bound provided by Lemma~\ref{lemma-surogate-representations}, we can also bound from above the error probability as stated in the following corollary: 
\begin{corollary}[Misclassification probability and information complexity] \label{corollary2}
Let $\mathcal{F}$ be a class of randomized encoders and decoders. Then, for every  $P_{XY}$ and every $\delta\in(0,1)$, with probability at least $1-\delta$ over the choice of $\mathcal{S}_n\sim P_{XY}^n$, the following inequalities hold $\forall\, Q=(Q_{U|X},Q_{\hat{Y}|U})\in\mathcal{F}$: 
\begin{align}
P_{\mathcal{E}}(Q)&\geq \mathcal{R}_{Y,Q_{\hat{Y}|U}}^{-1}\left( \mathcal{I}(\hat{P}_X; {Q}_{U|X})  \right)+ \mathcal{O}\left(\frac{1}{\sqrt{n}}\right), \\
P_{\mathcal{E}}(Q) &\leq 1 - \exp \left[- \mathcal{L}_{\text{emp}}(Q,{\mathcal{S}_n})- A_\delta\sqrt{\mathcal{I}(\hat{P}_X;Q_{U|X})}\cdot\frac{\log(n)}{\sqrt{n}} \right] +\mathcal{O}\left(\frac{1}{\sqrt{n}}\right).
\end{align}
\end{corollary}
It is worth to mention that the upper bound is an increasing function while the lower bound is a non-increasing function of the information complexity. Therefore, reducing the information complexity leads to tightening of both bounds.

\section{Information Complexity Case Studies}\label{Section-4}

The first part of this section is devoted to show that our investigated framework is general enough to encompass different learning frameworks. In particular, we study the information complexity of Generative Adversarial Networks (GANs)~\cite{NIPS2014_5423}. The second part looks at the role of dropout~\cite{srivastava2014dropout} that aims to directly control the information complexity. It is shown that there exists a non-trivial solution for the required amount of noise that optimizes the surrogate of the true risk~\eqref{eq:empirical_with_reg}.
\subsection{Generative Adversarial Networks}

Generative models try to learn to capture the statistical distribution of training data, allowing us to synthesize samples from the learned distribution. This is the key motivation behind GANs~\cite{NIPS2014_5423}.  Adversarial examples are a phenomenon where the neural network predictions can be significantly altered by adding seemingly imperceptible noise to an input. This phenomenon can be connected to the framework studied in this paper as follows.

In a basic GAN, the discriminator  is a binary classifier defined by $Q_{\hat{Y}|U}:\mathcal{U}\rightarrow \{0,1\}$ that randomly maps representations $U\in \mathcal{U}$ to binary labels $Y\in \mathcal{Y}\equiv\{0,1\}$, indicating if representations are from real samples or from a (fake) generator distribution. The generator distribution is denoted by $G_{X}\in\mathcal{P}(\mathcal{X})$. For a fixed generator $G_{X}$ the discriminator $Q_{\hat{Y}|U}$ is trained to classify data as either from training data or from fixed generator. When the discriminator is well trained, it may be frozen and the generator may be continue to be trained so as to lower the accuracy of the discriminator. The goal is that the generator distribution be able to match the real data distribution perfectly, i.e.,  $G_{X} \approx P_X$ so the discriminator will be maximally confused. Accordingly, we will refer to $P_{X}$ to the true data distribution and $P_Y\in \mathcal{P}(\mathcal{Y})$ to be the binary uniform distribution indicating if the encoder is providing representations that come from samples of the generator or from the training dataset. 

Training GANs involves both finding the discriminator that maximize its classification accuracy, and finding the parameters of a generator which maximally confuse the discriminator. The cost training is evaluated using the cross-entropy loss that depends on the generator and the discriminator. We can define our randomized  encoder as follows: 
\begin{equation}
Q_{U|XY}(u|xy)=\left\{
\begin{array}{lll}
\mathds{1}[u=x], & & \textrm{if $y=0$}\\
G_X(u), & & \textrm{if $y=1$}
\end{array}
\right.\label{eq-encoder-GAN}
\end{equation}
where the encoder inputs are given by $n$-data set $\mathcal{S}_n=\{(x_1,y_1)\cdots (x_n,y_n) \}$ according to ${P}_{XY} = {P}_X \cdot P_Y$. It is not difficult to check that the misclassification error~\eqref{eq-true-error} of our classifier  writes as:
\begin{align}
P_{\mathcal{E}}(Q_{\hat{Y}|U},Q_{{U}|XY}) &= 1- \sum\limits_{\forall\, (x,u,y) } Q_{{U}|XY}(u|xy) Q_{\hat{Y}|U}(y|u) P_X(x) P_Y(y)\\
&= \frac12 \mathbb{E}_{{P}_X} \big[ Q_{\hat{Y}|U} (y=0|X) \big] + \frac12 \mathbb{E}_{G_X} \big[ Q_{\hat{Y}|U} (y=1|X) \big].\nonumber
\end{align}

Notice that in this case the encoder is provided with the labels as well and produce representations according to the randomized encoder~\eqref{eq-encoder-GAN}. The empirical loss~\eqref{eq-EM-def} for the present model is written as:
\begin{align}
\mathcal{L}_{\textrm{GAN}}(Q_{\hat{Y}|U},Q_{U|XY}) &\coloneqq \mathbb{E}_{\hat{P}_{X}}  \mathbb{E}_{{P}_{Y}}  \big[ \ell\big({Q}_{U|XY}(\cdot|XY),Q_{\hat{Y}|U}(Y | \cdot)\big)  \big] \\
&= \frac12 \mathbb{E}_{\hat{P}_X} \big[ -\log Q_{\hat{Y}|U} (y=0|X) \big]+ \frac12 \mathbb{E}_{G_X} \big[ -\log \big( 1- Q_{\hat{Y}|U} (y=0|X) \big)\big],\label{loss-GAN}
\end{align}  
  
which is not difficult to verify that coincides with the original loss presented in~\cite{NIPS2014_5423}. Additionally, the information complexity of this model can be bounded for all $\tilde{Q}_U \in\mathcal{P}(\mathcal{U})$ by:
\begin{equation}
\mathcal{I}\big(\hat{P}_{XY};Q_{U|XY}\big) \leq   \mathcal{D}\left(Q_{U|XY} \big\| \tilde{Q}_U \big |\hat{P}_{XY} \right),\label{eq:information_radius}
 \end{equation}
where $\hat{P}_{XY}=\hat{P}_{X}{P}_{Y}$. Let us choose $\tilde{Q}_U\equiv G_X$ from which it is straightforward to show that:
\begin{equation}
\mathcal{I}\big(\hat{P}_{XY};Q_{U|XY}\big)\leq \frac{1}{2}\mathbb{E}_{\hat{P}_X} \big[ -\log G_X (X) \big],
\end{equation}
which induces an upper bound to the information complexity given by the cross-entropy loss for GAN  and yields the following penalized risk: 
\begin{IEEEeqnarray}{rCl}
  \mathcal{L}_{\textrm{GAN}}(Q_{\hat{Y}|U},Q_{U|XY})  + \lambda\cdot   \sqrt{\mathbb{E}_{\hat{P}_X} \big[ -\log G_X (X) \big]},     \label{eq-GAN-penality-final}
 \end{IEEEeqnarray}
 where $\lambda>0$ is a suitable multiplier. Despite the theoretical solutions GANs training is still challenging and often unstable for several reasons~\cite{DBLP:journals/corr/ArjovskyB17}. Can this penalty suggest a way to defend against the discriminator loss to converge to zero, which provides no reliable path for gradient updates to the generator? The answer to this question deserves further studies beyond the scope of this paper.

\subsection{Dropout and Information Complexity}\label{section-dropout-complex}

We will investigate surrogate bounds to the information complexity of the dropout models~\cite{srivastava2014dropout} on two differents networks: (1) Restricted Boltzmann Machines (RBMs) and (2) a Feed-Forward Network. We will show explicitly how the dropout parameter $p_{\rm out}$ controls the mutual information term $\mathcal{I}\left(P_{X};Q_{U|X}\right)$ and   $\mathcal{L}(Q_{\hat{Y}|U},Q_{U|X})$. Moreover, it is also pointed out that in general there exists an optimal value of $p_{\rm out}\in(0,1)$ which minimizes the objective: $\mathcal{L}(Q_{\hat{Y}|U},Q_{U|X})+\lambda\cdot\sqrt{\mathcal{I}\left(P_{X};Q_{U|X}\right)}$. Our analysis and conclusions here apply to a generic input distribution $P_{XY}$ such as the empirical distribution $\hat{P}_{XY}$. This yields an upper bound to the information complexity and indicates that the results of this section could have practical implications. Although the analysis for simplicity is done for a network with one hidden layer, our results can be used to bound the information complexity of deep networks by resorting to the bound (\ref{eq-bound-multilayer}).

\subsubsection{Restricted Boltzmann Machines}
We begin by expressing the distribution of the hidden units given the input ones and the noise random variables~\cite{srivastava2014dropout}:
\begin{equation}
Q_{\mathbf{U}|\mathbf{X}\mathbf{Z}}(\mathbf{u}|\mathbf{x},\mathbf{z})=\prod_{i=1}^mQ_{U_i|\mathbf{X}Z_i}(u_i|\mathbf{x},z_i),
\label{eq:RBM_hidden}
\end{equation}  
where $\mathbf{u}\in\mathcal{U}\equiv \{0,1\}^{m}$ is the vector of hidden units , $\mathbf{x}\in\mathcal{X}\equiv \{0,1\}^{d}$ is the vector of input (or visible units) and $\mathbf{z}\in\mathcal{Z}\equiv \{0,1\}^m$ is the noise vector\footnote{Note, in contrast with previous sections, we explicitly take into account the vectorial nature of the variables $\mathbf{U}$, $\mathbf{X}$ and $\mathbf{Z}$. This is done in order to explicitly include in the results the dependence with the dimensions of the visible and hidden units.}. It is not difficult to show that, for $i=1,\dots,m$:
\begin{equation*}
Q_{U_i|\mathbf{X}Z_i}(u_i|\mathbf{x},z_i=1)=\left\{\begin{array}{ll}
\sigma(b_i+\langle\mathbf{w}_i,\mathbf{x}\rangle), & u_i=1\\
1- \sigma(b_i+\langle\mathbf{w}_i,\mathbf{x}\rangle), & u_i=0
\end{array}\right.
\end{equation*}
\begin{equation}
Q_{U_i|\mathbf{X}Z_i}(u_i|\mathbf{x},z_i=0)=\left\{\begin{array}{cc}
0, & u_i=1\\
1, & u_i=0
\end{array}\right.
\label{eq:Q_marg_RBM_2}
\end{equation}
where $\sigma(x)\coloneqq\frac{1}{1+\exp{(-x)}}$ and where $(\mathbf{w}_i,b_i)$ are the parameters corresponding to the hidden unit $i$ with $i=1,\dots,m$. Using the fact that $P_{\mathbf{Z}}(\mathbf{z})=\prod_{i=1}^{m}p_{\rm out}^{z_i}(1-p_{\rm out})^{1-z_i}$, $p_{\rm out}\in[0,1]$, we can write:
\begin{equation}
Q_{\mathbf{U}\mathbf{Z}|\mathbf{X}}(\mathbf{u},\mathbf{z}|\mathbf{x})=\prod_{i=1}^m
Q_{U_i|\mathbf{X}Z_i}(u_i|\mathbf{x},z_i)p_{\rm out}^{z_i}(1-p_{\rm out})^{1-z_i}.
\end{equation}
Marginalizing with respect to $\mathbf{z}$ we obtain:
\begin{equation}
Q_{\mathbf{U}|\mathbf{X}}(\mathbf{u}|\mathbf{x})=\prod_{i=1}^m
Q_{U_i|\mathbf{X}}(u_i|\mathbf{x}),
\label{eq:Q_marg_RBM_3}
\end{equation}
where
\begin{align}
Q_{U_i|\mathbf{X}}(u_i|\mathbf{x})=&\mathds{1}\left[u_i=1\right]p_{\rm out}\sigma(b_i+\langle\mathbf{w}_i,\mathbf{x}\rangle)+\mathds{1}\left[u_i=0\right]\left[1-p_{\rm out} \sigma(b_i+\langle\mathbf{w}_i,\mathbf{x}\rangle)\right].
\label{eq:Q_marg_RBM_4}
\end{align}
We see that $U_i$ given $\mathbf{X}$ is distributed as a Bernoulli random variable with parameter $p_{\rm out}\sigma(b_i+\langle\mathbf{w}_i,\mathbf{X}\rangle)$. We are interested in computing $\mathcal{I}\left(P_{\mathbf{X}};Q_{\mathbf{U}|\mathbf{X}}\right)$. This is hard to compute quantity. For this reason, we will use the variational characterization of mutual information (\ref{eq:information_radius}), which can be written as:
\begin{equation}
\mathcal{I}\left(P_{\mathbf{X}};Q_{\mathbf{U}| \mathbf{X}}\right) = \min_{\tilde{Q}_{\mathbf{U}} \in\mathcal{P}(\mathcal{U})}  \mathcal{D}\left(Q_{\mathbf{U}|\mathbf{X}} \big\| \tilde{Q}_{\mathbf{U}}\big |P_{\mathbf{X}} \right).
\end{equation}
We will use this characterization to find an optimized upper-bound of mutual information in terms of an easily to compute divergence term. Consider a product distribution $\tilde{Q}_{\mathbf{U}}(\mathbf{u})=\prod_{i=1}^{n}\tilde{Q}_{U_i}(u_i)$. It is straightforward to see that:
\begin{align}
\mathcal{I}\left(P_{\mathbf{X}};Q_{\mathbf{U}| \mathbf{X}}\right) \leq \min_{\tilde{Q}_{\mathbf{U}}=\prod_{i=1}^{n}\tilde{Q}_{U_i}(u_i)} \mathbb{E}_{P_{\mathbf{X}}} \left[\sum_{i=1}^{m} \mathcal{D}\left(Q_{U_i|\mathbf{X}}(\cdot|\mathbf{X}) \big\| \tilde{Q}_{U_i} \right)\right].
\label{eq:inf_radius_opt}
\end{align}
The optimization problem in (\ref{eq:inf_radius_opt}) can be solved in closed form. We have the following lemma:
\begin{lemma}[Optimal product distribution]\label{lem:qopt}
Consider the optimization problem in (\ref{eq:inf_radius_opt}). The optimal product distribution $\tilde{Q}_{\mathbf{U}}(\mathbf{u})=\prod_{i=1}^{n}\tilde{Q}_{U_i}(u_i)$ is given by:
$
\tilde{Q}_{U_i}(u_i)=\mathbb{E}_{P_{\mathbf{X}}} \left[Q_{U_i|\mathbf{X}}(\cdot|\mathbf{X})\right],
$ $i=[1:m]$.
\end{lemma} 
\begin{IEEEproof}
The proof follows easily from the non-negativity of the divergence and from  the following observation, for every $i=[1:m]$:
\begin{align}
 \mathcal{D}\left(Q_{U_i|\mathbf{X}} \big\| \tilde{Q}_{U_i} \big| P_{\mathbf{X}} \right)&= \mathcal{D}\left(Q_{U_i|\mathbf{X}}  \big\| \mathbb{E}_{P_{\mathbf{X}}} \left[Q_{U_i|\mathbf{X}}(\cdot|\mathbf{X}) \right] \big| P_{\mathbf{X}} \right) +\mathcal{D}\left(\mathbb{E}_{P_{\mathbf{X}}} \left[Q_{U_i|\mathbf{X}}(\cdot|\mathbf{X})\right] \big\| \tilde{Q}_{U_i} \right).
\end{align}	
\end{IEEEproof}

From the above lemma we see that the optimal distribution for each $i=[1:m]$ is the marginalization of $Q_{U_i|\mathbf{X}}(\cdot|\mathbf{X})$ w.r.t. $\mathbf{X}$. We also obtain the following bound for mutual information:
\begin{equation}
\mathcal{I}\left(P_{\mathbf{X}};Q_{\mathbf{U}| \mathbf{X}}\right) \leq\sum_{i=1}^m \mathcal{D}\left(Q_{U_i|\mathbf{X}} \big\| \mathbb{E}_{P_{\mathbf{X}}} \left[Q_{U_i|\mathbf{X}}(\cdot|\mathbf{X})\right] \big| P_{\mathbf{X}}\right).
\label{eq:bound_mutual_inf_radius}
\end{equation}
From (\ref{eq:Q_marg_RBM_4}) we can write:
\begin{IEEEeqnarray}{CCl}
&\mathbb{E}_{P_{\mathbf{X}}}&\left[Q_{U_i|\mathbf{X}}(u_i|\mathbf{x})\right]=\mathds{1}\left[u_i=1\right]p_{\rm out} \mathbb{E}_{P_{\mathbf{X}}}\left[\sigma(b_i+\langle\mathbf{w}_i,\mathbf{x}\rangle)\right] +\mathds{1}\left[u_i=0\right]\left[1-p_{\rm out} \mathbb{E}_{P_{\mathbf{X}}}\left[\sigma(b_i+\langle\mathbf{w}_i,\mathbf{x}\rangle)\right]\right].
\end{IEEEeqnarray}
Finally, observing that $\mathbb{E}_{P_{\mathbf{X}}}\left[Q_{U_i|\mathbf{X}}(u_i|\mathbf{x})\right]$ and $Q_{U_i|\mathbf{X}}(u_i|\mathbf{x})$ are the distribution of \emph{Bernoulli} RVs, we can write:
\begin{align}
\mathcal{I}\left(P_{\mathbf{X}};Q_{\mathbf{U}| \mathbf{X}}\right)& \leq \sum_{i=1}^m\mathbb{E}_{P_{\mathbf{X}}} \left[p_{\rm out} \sigma(b_i+\langle\mathbf{w}_i,\mathbf{x}\rangle)\log{\left(\frac{\sigma(b_i+\langle\mathbf{w}_i,\mathbf{x}\rangle)}{\mathbb{E}_{P_{\mathbf{X}}}\left[\sigma(b_i+\langle\mathbf{w}_i,\mathbf{x}\rangle)\right]}\right)}  \right.\nonumber\\
&+ \left.\left[1-p_{\rm out}\sigma(b_i+\langle\mathbf{w}_i,\mathbf{x}\rangle)\right]\log{\left(\frac{1-p_{\rm out}\sigma(b_i+\langle\mathbf{w}_i,\mathbf{x}\rangle)}{1-p_{\rm out}\mathbb{E}_{P_{\mathbf{X}}}\left[\sigma(b_i+\langle\mathbf{w}_i,\mathbf{x}\rangle)\right]}\right)}\right].\label{eq:bound_mutual_inf_radius_2}
\end{align}

Similarly, $\mathcal{L}(Q_{\hat{Y}|\mathbf{U}},Q_{\mathbf{U}|\mathbf{X}})\equiv\mathcal{L}(Q)$ can be written as:
\begin{align}
\mathcal{L}(Q)&=\mathbb{E}_{P_{\mathbf{XY}}}\left[-\sum_{\forall\,\mathbf{u}}Q_{\mathbf{U}|\mathbf{X}}(\mathbf{u}|\mathbf{X})\log{Q_{\hat{Y}|\mathbf{U}}(Y|\mathbf{u})}\right]\\
&=\sum_{\forall\,\mathbf{u}}\mathbb{E}_{P_{\mathbf{X}}}\left[\mathbb{E}_{P_{\mathbf{Y|X}}}\left[-\log{Q_{\hat{Y}|\mathbf{U}}(Y|\mathbf{u})}\right]\prod_{i=1}^m
\left[p_{\rm out}\sigma(b_i+\langle\mathbf{w}_i,\mathbf{x}\rangle)\right]^{u_i}
\left[1-p_{\rm out}\sigma(b_i+\langle\mathbf{w}_i,\mathbf{x}\rangle)\right]^{1-u_i}\right].\label{eq:log_loss_RBM}
\end{align}
Defining $\lambda\geq 0$, using (\ref{eq:log_loss_RBM}) and taking the square-root of the right-side term in (\ref{eq:bound_mutual_inf_radius_2}) we can obtain an upper bound on $\mathcal{L}(Q_{\hat{Y}|\mathbf{U}},Q_{\mathbf{U}|\mathbf{X}})+\lambda\sqrt{\mathcal{I}\left(P_{\mathbf{X}};Q_{\mathbf{U}| \mathbf{X}}\right)}$ from which we can proceed with our analysis. That is, following the conventions for the parametrization of the encoder and decoder given in section \ref{subsec:restricted}, we will consider the following cost function:
\begin{equation}
\mathcal{J}(\bm{\theta},\bm{\gamma},p_{\rm out})\coloneqq \mathcal{L}\left(Q_{\hat{Y}|\mathbf{U}}^{(\bm{\gamma})},Q_{\mathbf{U}|\mathbf{X}}^{(\bm{\theta})}\right)+\lambda\sqrt{\mathcal{I}\left(P_{\mathbf{X}};Q_{\mathbf{U}|\mathbf{X}}^{(\bm{\theta})}\right)},\label{eq:cost_C}
\end{equation}
where $\bm{\theta}=\left\{(b_i,\mathbf{w}_i)\right\}_{i=1}^{m}$ are the parameters of the encoder $Q_{\mathbf{U}|\mathbf{X}}^{(\bm{\theta})}$ and $\bm{\gamma}$ are the parameters of the decoder $Q_{\hat{Y}|\mathbf{U}}^{(\bm{\gamma})}$. Let us assume that the decoder is given by the \emph{soft-max} function\footnote{We assume implicitly that parameters are given such that the misclassification probability is lower than the random classifier.} which is a natural choice when the number of classes is $|\mathcal{Y}|>2$. In this way:
\begin{equation}
Q_{\hat{Y}|\mathbf{U}}^{(\bm{\gamma})}(y=j|\mathbf{u})=\frac{\exp{\langle\mathbf{a}_j,\mathbf{u}\rangle}}{\sum\limits_{k=1}^{|\mathcal{Y}|}\exp{\langle\mathbf{a}_k,\mathbf{u}\rangle}},\ \ j=1,\dots,|\mathcal{Y}|,
\label{eq:softmax}
\end{equation}
where $\bm{\gamma}=\left\{\mathbf{a}_i\right\}_{i=1}^{|\mathcal{Y}|}$ are the parameters of the decoder. To keep the notation simpler we will define:
\begin{equation}
f(\bm{\gamma},\mathbf{x},\mathbf{u})\coloneqq \mathbb{E}_{P_{\mathbf{Y|X}}}\left[-\log{Q_{\hat{Y}|\mathbf{U}}^{(\bm{\gamma})}(Y|\mathbf{u})}\right].
\label{eq:f_def}
\end{equation}
We will assume that the parameters $\bm{\gamma}$ are constrained to belong to a set $\Gamma$ such that:
\begin{equation}
f(\bm{\gamma},\mathbf{x},\mathbf{u})\leq f(\bm{\gamma},\mathbf{x},\mathbf{0})=\log {|\mathcal{Y}|},
\label{eq:assumption_f}
\end{equation}
for all $(\bm{\gamma},\mathbf{x},\mathbf{u})\in\Gamma\times\{0,1\}^m\times\{0,1\}^d$. That is, the log-loss of the decoders that can be implemented cannot be worse (on average conditioned on the input $\mathbf{x}$) than $\log{|\mathcal{Y}|}$, which is the value of the log-loss of the uniform decoder which select the class to which belongs its input uniformly in $\mathcal{Y}$. 

The parameters $(\bm{\theta},\bm{\gamma})$ should be chosen such that $\mathcal{J}(\bm{\theta},\bm{\gamma},p_{\rm out})$ is minimized. We see that this cost function depends on the dropout parameter $p_{\rm out}\in[0,1]$ which can be thought as a hyperparameter. We can analyze the dependence of $\mathcal{J}(\bm{\theta},\bm{\gamma},p_{\rm out})$ with $p_{\rm out}$ to see if there is an optimal non-trivial choice for this quantity. The answer is affirmative and it is summarized in the next lemma.

\begin{lemma}\label{lemma-result-Pout}
Consider $\mathcal{J}(\bm{\theta},\bm{\gamma},p_{\rm out})$ defined as in (\ref{eq:cost_C}) and a single layer RBM network given by the family of encoders in (\ref{eq:Q_marg_RBM_3}) and (\ref{eq:Q_marg_RBM_4}), and the family of decoders in (\ref{eq:softmax}). Assume also that condition (\ref{eq:assumption_f}) is fulfilled. Then, for every set of parameters $(\bm{\theta},\bm{\gamma})$ the optimal value $p_{\rm out}$ (i.e. $\mathcal{J}(\bm{\theta},\bm{\gamma},p_{\rm out})\leq \mathcal{J}(\bm{\theta},\bm{\gamma},p_{\rm out} )$ for all $p_{\rm out} \in[0,1]$) is non-trivial. That is, $p_{\rm out} \in(0,1)$.
\end{lemma}
\begin{IEEEproof}
Consider each term in the sum in the right-hand side of (\ref{eq:bound_mutual_inf_radius_2}). In order to simplify notation we define for each $i=[1:m]$:
\begin{align}
g_i(p,b_i,\mathbf{w}_i)\coloneqq &\mathbb{E}_{P_{\mathbf{X}}} \left[p_{\rm out}  \sigma(b_i+\langle\mathbf{w}_i,\mathbf{x}\rangle)\log{\left(\frac{\sigma(b_i+\langle\mathbf{w}_i,\mathbf{x}\rangle)}{\mathbb{E}_{P_{\mathbf{X}}}\left[\sigma(b_i+\langle\mathbf{w}_i,\mathbf{x}\rangle)\right]}\right)}+\right.\nonumber\\
&\left.\left[1-p_{\rm out} \sigma(b_i+\langle\mathbf{w}_i,\mathbf{x}\rangle)\right]\log{\left(\frac{1-p_{\rm out} \sigma(b_i+\langle\mathbf{w}_i,\mathbf{x}\rangle)}{1-p_{\rm out} \mathbb{E}_{P_{\mathbf{X}}}\left[\sigma(b_i+\langle\mathbf{w}_i,\mathbf{x}\rangle)\right]}\right)}\right].
\end{align}
We can calculate the derivative of these terms with respect to $p_{\rm out} $ obtaining:
\begin{align}
\frac{\partial g_i(p,b_i,\mathbf{w}_i) }{\partial p_{\rm out} }=&\mathbb{E}_{P_{\mathbf{X}}} \left[\sigma(b_i+\langle\mathbf{w}_i,\mathbf{x}\rangle)\log{\left(\frac{\sigma(b_i+\langle\mathbf{w}_i,\mathbf{x}\rangle)}{\mathbb{E}_{P_{\mathbf{X}}}\left[\sigma(b_i+\langle\mathbf{w}_i,\mathbf{x}\rangle)\right]}\right)}\right]\nonumber\\
& -\mathbb{E}_{P_{\mathbf{X}}} \left[\sigma(b_i+\langle\mathbf{w}_i,\mathbf{x}\rangle)\log{\left(\frac{1-p_{\rm out} \sigma(b_i+\langle\mathbf{w}_i,\mathbf{x}\rangle)}{1-p_{\rm out} \mathbb{E}_{P_{\mathbf{X}}}\left[\sigma(b_i+\langle\mathbf{w}_i,\mathbf{x}\rangle)\right]}\right)}\right].
\end{align}

Using the fact that $x\log{x}$ is strictly convex for every $x\in\mathbb{R}_{\geq 0}$ and that $x\log{(1-p_{\rm out} x)}$ is strictly concave for each $x\in[0,1]$ and $p_{\rm out} \in(0,1]$ we can use Jensen inequality to show that\footnote{In order to have the strict inequality in (\ref{eq:strict_ineq}) we also need that $\mathbf{X}$ be a non-degenerate random vector. It is clear that this would be the situation in practice.}:
\begin{equation}
\frac{\partial g_i(p,b_i,\mathbf{w}_i) }{\partial p_{\rm out} }>0,
\label{eq:strict_ineq}
\end{equation}
for every $(p_{\rm out} ,b_i,\mathbf{w}_i)$. This implies that $g_i(p_{\rm out} ,b_i,\mathbf{w}_i)$ is strictly increasing with $p_{\rm out} \in[0,1]$ for every $i=[1:m]$, which also implies that the second term in (\ref{eq:cost_C}) is strictly increasing with $p_{\rm out} \in[0,1]$. Similarly consider $\mathcal{L}\big(Q_{\hat{Y}|\mathbf{U}}^{(\bm{\gamma})},Q_{\mathbf{U}|\mathbf{X}}^{(\bm{\theta})}\big)$ in (\ref{eq:log_loss_RBM}) relabelled as $s(\bm{\theta},\bm{\gamma},p_{\rm out} )$:
\begin{align}
s(\bm{\bm{\gamma},\theta},p_{\rm out} )&\coloneqq\sum_{\forall\,\mathbf{u}}\mathbb{E}_{P_{\mathbf{X}}}\left[f(\bm{\gamma},\mathbf{x},\mathbf{u})\prod_{i=1}^m
\left[p_{\rm out}  \sigma(b_i+\langle\mathbf{w}_i,\mathbf{x}\rangle)\right]^{u_i}
\left[1-p_{\rm out} \sigma(b_i+\langle\mathbf{w}_i,\mathbf{x}\rangle)\right]^{1-u_i}\right].
\end{align}
From condition (\ref{eq:assumption_f}) it is clear that:
\begin{equation}
s(\bm{\theta},\bm{\gamma},p_{\rm out} )\leq \log{|\mathcal{Y}|},\ \forall\,(\bm{\gamma},\bm{ \theta},p_{\rm out} )\in\Gamma\times\Theta\times[0,1].
\end{equation}
But it is easy to show that $s(\bm{\theta},\bm{\gamma},0)=\log{|\mathcal{Y}|}$ and $\frac{\partial s(\bm{\theta},\bm{\gamma},p_{\rm out} )}{\partial p_{\rm out} }\Big|_{p_{\rm out} =0}<0$. This implies that, at least in a vicinity of $p_{\rm out} =0$, $s(\bm{\theta},\bm{\gamma},p_{\rm out} )$ is decreasing in $p_{\rm out} $. This concludes the proof of the lemma.
\end{IEEEproof}

\begin{remark}
The  consequence of Lemma~\ref{lemma-result-Pout} is that there is a non-trivial solution, i.e., $p_{\rm out}\neq \{0,1\}$, to the optimization problem in~\eqref{eq-principle-IT} of jointly minimizing the average loss function and the information complexity. Although the determination of the optimal $p_{\rm out}$ does not appear to be feasible, this result states that the presence of noise (randomization) during training is not only useful to enforce invariant representations but also can help to minimize the surrogate  of the overall risk~\eqref{eq:tradeoff-empgen}. 
\end{remark}



\subsubsection{Feed-forward neural networks with logistic activation}
\label{subsec:feedforward}
Following with the previous notation, in this model  hidden units can be expressed as a deterministic function of input and dropout variable (see~\cite{srivastava2014dropout}):  $u_i=\sigma(b_i+\langle\mathbf{w}_i,\mathbf{x}\circledcirc\mathbf{z}\rangle)$, where $\circledcirc$ denotes the Hadamard product. Given the input units $\mathbf{x}$, the sigmoid argument is a sum of independent Bernoulli random variables $\{Z_k\}_{k=1}^d$ and the \emph{Central limit Theorem} (CLT) can be used to obtain \cite{Wang_Manning_2013}:
\begin{equation}
b_i+\langle\mathbf{w}_i,\mathbf{x}\circledcirc\mathbf{Z}\rangle\sim\mathcal{N}\left(b_i+p_{\rm out}\langle\mathbf{w}_i,\mathbf{x}\rangle;p_{\rm out}(1-p_{\rm out})\langle\mathbf{x}\circledcirc\mathbf{w}_i,\mathbf{x}\circledcirc\mathbf{w}_i\rangle\right).
\end{equation} 
In order to simplify the analysis we approximate $\sigma(x)=\mathds{1}[x>0]$ \footnote{This allows to consider $\mathbf{u}\in\{0,1\}^m$ when in a feed-forward neural net $\mathbf{u}\in[0,1]^m$, although strictly speaking the actual support of $\mathcal{U}$ is discrete because $\mathcal{X}$ and $\mathcal{Z}$ are discrete.} to express the distribution of each hidden unit given the input ones as:
\begin{equation}
Q_{U_i|\mathbf{X}}(1|\mathbf{x})\approx\Phi\left(\frac{b_i+p_{\rm out}\langle\mathbf{w}_i,\mathbf{x}\rangle}{\sqrt{p_{\rm out}(1-p_{\rm out})\langle\mathbf{x}\circledcirc\mathbf{w}_i,\mathbf{x}\circledcirc\mathbf{w}_i\rangle}}\right),
\end{equation}
where $\Phi(x)=\frac{1}{\sqrt{2\pi}}\int_{x}^\infty e^{-\frac{t^2}{2}}dt$. Also, we simplify the model assuming conditional independence as:
\begin{equation}
Q_{\mathbf{U}|\mathbf{X}}(\mathbf{u}|\mathbf{x})\approx\prod_{i=1}^mQ_{U_i|\mathbf{X}}(1|\mathbf{x})^{u_i}\left(1-Q_{U_i|\mathbf{X}}(1|\mathbf{x})\right)^{1-u_i}.
\label{eq:product_dist_gaussian_assump}
\end{equation}
As it was done previously we can use the mutual information upper bound given by \eqref{eq:bound_mutual_inf_radius}. From (\ref{eq:product_dist_gaussian_assump}) it is easy to compute each relative entropy  term (as the involved distributions are \emph{Bernoulli}). In order to save space we do not show the final expression. In Section~\ref{Section-6}, we will use this result considering a feed-forward network with several layers. In order to do that, besides the results of this section, we will also use \eqref{eq-bound-multilayer}.

\section{Proof of Theorem~\ref{thm:regularizar}} \label{sec:proof_main_theorem}

In this section, we present the proof of Theorem~\ref{thm:regularizar} for which we shall make use of some auxiliary results provided in the Appendices.

\begin{IEEEproof}
By letting $Q=(Q_{\hat{Y}|U},Q_{U|X})$, it is immediate to see that:
\begin{equation}
\label{eq:cost}
\mathcal{L}(Q)=\sum_{(u,y)\in\mathcal{U}\times\mathcal{Y}}Q_{UY}(u,y)\log{\left(\frac{1}{Q_{\hat{Y}|U}(y|u)}\right)},
\end{equation}
\begin{equation}
\label{eq:cost_emp}
\mathcal{L}_{\rm emp}(Q,\mathcal{S}_n)=\sum_{(u,y)\in\mathcal{U}\times\mathcal{Y}}\hat{Q}_{UY}(u,y)\log{\left(\frac{1}{Q_{\hat{Y}|U}(y|u)}\right)},
\end{equation}
where:
\begin{IEEEeqnarray}{rCl}
Q_{UY}(u,y)&=&\sum_{x\in\mathcal{X}}Q_{U|X}(u|x)P_{XY}(x,y),\\
\hat{Q}_{UY}(u,y)&=&\sum_{x\in\mathcal{X}}Q_{U|X}(u|x)\hat{P}_{XY}(x,y).
\end{IEEEeqnarray}
These expressions allow us to write:  
\begin{align}
\mathcal{E}_{\textrm{gap}}(Q,\mathcal{S}_n)=&\left|\sum_{\forall (u,y)}\left[Q_{YU}(y,u)-\hat{Q}_{YU}(y,u)\right]\log\left(\frac{1}{Q_{\hat{Y}|U}(y|u)}\right)\right|\nonumber\\
\leq&\left|\sum_{\forall (u,y)}\left[Q_{YU}(y,u)-\hat{Q}_{YU}(y,u)\right]\log\left(\frac{Q_{Y|U}(y|u)}{Q_{\hat{Y}|U}(y|u)}\right)\right|\nonumber\\
&+\left|\sum_{\forall (u,y)}\left[Q_{YU}(y,u)-\hat{Q}_{YU}(y,u)\right]\log\left(\frac{1}{Q_{Y|U}(y|u)}\right)\right|\nonumber\\
=&\left|\sum_{\forall (u,y)}\left[Q_{YU}(y,u)-\hat{Q}_{YU}(y,u)\right]\log\left(\frac{Q_{Y|U}(y|u)}{Q_{\hat{Y}|U}(y|u)}\right)\right|+\mathcal{E}_{\textrm{gap}}(Q_{Y|U},Q_{U|X},\mathcal{S}_n).\label{eq:primeracota}
\end{align}
The second term in $\eqref{eq:primeracota}$ depends only on the encoder $Q_{U|X}$, while the first term depends also on the decoder $Q_{\hat{Y}|U}$. The first term can be written as:
\begin{align}
\left|\sum_{\forall (u,y)}\left[Q_{YU}(y,u)-\hat{Q}_{YU}(y,u)\right]\log\left(\frac{Q_{Y|U}(y|u)}{Q_{\hat{Y}|U}(y|u)}\right)\right|=&\left|\sum_{\forall (x,y)}\left[P_{XY}(x,y)-\hat{P}_{XY}(x,y)\right]\right.\nonumber\\
&\left.\times\sum_{u\in\mathcal{U}}Q_{U|X}(u|x)\log\left(\frac{Q_{Y|U}(y|u)}{Q_{\hat{Y}|U}(y|u)}\right)\right|,\\
=&\left|\sum_{\forall (x,y)}\left[P_{XY}(x,y)-\hat{P}_{XY}(x,y)\right]\right.\nonumber\\
&\left.\times\left[\sum_{u\in\mathcal{U}}Q_{U|X}(u|x)\log\left(\frac{Q_{Y|U}(y|u)}{Q_{\hat{Y}|U}(y|u)}\right)-a\right]\right|,
\label{eq:prevariance}
\end{align}
where $a\in\mathbb{R}$ is arbitrary number. Inspired by~\cite{Shamir:2010:LGI:1808343.1808503}, we consider the definition:
\begin{equation}
\mathbb{V}(\mathbf{a})\coloneqq \left\|\mathbf{a}-\bar{a}\mathds{1}_d\right\|_2^2,
\label{eq:emp_var}
\end{equation}
where $\mathbf{a}\in\mathbb{R}^d$ with $d\in\mathbb{N}_+$, $\bar{a}=\frac{1}{d}\sum_{i=1}^d a_i$, and $\mathds{1}_d$ is the vector of ones of length $d$. This quantity measures the dispersion of the components of vector around their mean. Note that:
\begin{equation}
\mathbb{V}(\mathbf{a}) \leq\left\|\mathbf{a}-b\mathds{1}_d\right\|_2^2\ , \ \forall b\in\mathbb{R}.
\end{equation}
So setting $a\coloneqq\frac{1}{|\mathcal{X}||\mathcal{Y}|}\sum\limits_{\forall (x,y,u)}Q_{U|X}(u|x)\log\left(\frac{Q_{Y|U}(y|u)}{Q_{\hat{Y}|U}(y|u)}\right)$ and applying Cauchy-Schwartz inequality in \eqref{eq:prevariance} we obtain:
\begin{align}
&\left|\sum_{\forall (u,y)}\left[Q_{YU}(y,u)-\hat{Q}_{YU}(y,u)\right]\log\left(\frac{Q_{Y|U}(y|u)}{Q_{\hat{Y}|U}(y|u)}\right)\right|\leq \left\|\mathbf{P}_{XY}-\mathbf{\hat{P}}_{XY}\right\|_2{\Lambda}\big(Q_{U|X},{Q}_{Y|U},Q_{\hat{Y}|U}\big),\,\,\,
	\end{align}
where we have denoted: 
\begin{align}
{\Lambda}\big(Q_{U|X},{Q}_{Y|U},Q_{\hat{Y}|U}\big)\coloneqq\sqrt{\mathbb{V}\left( \left\{\sum_{u\in\mathcal{U}}Q_{U|X}(u|{x})\log\left(\frac{Q_{Y|U}({y}|u)}{Q_{\hat{Y}|U}({y}|u)}\right)\right\}_{(x,y)\in\mathcal{X}\times\mathcal{Y}}\right)}.
\end{align}

The second term of \eqref{eq:primeracota} can be bounded as:
\begin{align}
\mathcal{E}_{\textrm{gap}}(Q_{Y|U},Q_{U|X},\mathcal{S}_n) &=\left|\sum_{\forall(u,y)}\left[Q_{YU}(y,u)-\hat{Q}_{YU}(y,u)\right]\log\left(\frac{1}{Q_{Y|U}(y|u)}\right)\right|\nonumber\\
&\leq\left| \mathcal{H}(Q_{Y|U}|Q_U) - \mathcal{H}(\hat{Q}_{Y|U}|\hat{Q}_U)\right|+ \mathcal{D}\left(\hat{Q}_{Y|U}\big\|Q_{Y|U}\big| \hat{Q}_U \right).\label{eq:segundacota}
\end{align}
The first term of \eqref{eq:segundacota} can be bounded as:
\begin{align}
\left| \mathcal{H}(Q_{Y|U}|Q_U)- \mathcal{H}(\hat{Q}_{Y|U}|\hat{Q}_U)\right|&\leq\left| \mathcal{H}(P_Y)- \mathcal{H}(\hat{P}_{Y})\right|\nonumber\\
&+\left| \mathcal{H}(Q_{U|Y}|P_Y)- \mathcal{H}(\hat{Q}_{U|Y}|\hat{P}_Y)\right|+\left| \mathcal{H}(Q_{U})- \mathcal{H}(\hat{Q}_{U})\right|.
\end{align}
We bound the second term of \eqref{eq:segundacota} with the standard Kullback-Leibler inequalities: 
\begin{align}
\mathcal{D}\left(\hat{Q}_{Y|U}\big\|Q_{Y|U}\big| \hat{Q}_U \right)&\leq\mathcal{D}\left(\hat{Q}_{UY}\big\|Q_{UY}\right)\leq\mathcal{D}\left(\hat{Q}_{UXY}\big\|Q_{UXY}\right)=\mathcal{D}\left(\hat{P}_{XY}\big\|P_{XY}\right).
\end{align}
As a consequence, the generalization gap can be bounded using Lemmas~\ref{lem:Hu} and~\ref{lem:Hu|y}:
\begin{align}
\mathcal{E}_{\textrm{gap}}(Q,\mathcal{S}_n)&\leq\left\|\mathbf{P}_{XY}-\mathbf{\hat{P}}_{XY}\right\|_2{\Lambda}\big(Q_{U|X},{Q}_{Y|U},Q_{\hat{Y}|U}\big) +\left| \mathcal{H}(Q_{U|Y}|P_Y)- \mathcal{H}(\hat{Q}_{U|Y}|\hat{P}_Y)\right|\nonumber\\
&+\left| \mathcal{H}(Q_{U})- \mathcal{H}(\hat{Q}_{U})\right| +\left| \mathcal{H}(P_Y)- \mathcal{H}(\hat{P}_{Y})\right|+\mathcal{D}\left(\hat{P}_{XY}\|P_{XY}\right)\\
&\leq \left\|\mathbf{P}_{XY}-\mathbf{\hat{P}}_{XY}\right\|_2{\Lambda}\big(Q_{U|X},{Q}_{Y|U},Q_{\hat{Y}|U}\big)+\left| \mathcal{H}(P_Y)- \mathcal{H}(\hat{P}_{Y})\right|\nonumber\\
&+\mathcal{D}\left(\hat{P}_{XY}\big \|P_{XY}\right)+\sum_{u\in\mathcal{U}}\phi\left(\left\|\mathbf{P}_{X}-\mathbf{\hat{P}}_{X}\right\|_2\cdot\sqrt{\mathbb{V}(\{Q_{U|X}(u|{x})\}_{x\in\mathcal{X}} )}\right)\nonumber\\
&+\sum_{y\in\mathcal{Y}}P_Y(y)\sum_{u\in\mathcal{U}}\phi\left(\left\|\mathbf{P}_{X|Y}(\cdot|y)-\mathbf{\hat{P}}_{X|Y}(\cdot|y)\right\|_2\times\sqrt{\mathbb{V}(\{Q_{U|X}(u|{x})\}_{x\in\mathcal{X}})}\right),
\end{align}
where $\phi(\cdot)$ is defined in \eqref{eq:cotaphi}. It is clear that
$P_Y\mapsto  \mathcal{H}(P_Y)$ is a differentiable function and thus, we can apply a first order Taylor expansion to obtain:
\begin{align}
\mathcal{H}(P_Y)- \mathcal{H}(\hat{P}_Y) =\left\langle\frac{\partial  \mathcal{H}(P_Y)}{\partial \mathbf{P}_{Y}},\mathbf{P}_{Y}-\mathbf{\hat{P}}_{Y}\right\rangle+o\left(\|\mathbf{P}_{Y}-\mathbf{\hat{P}}_{Y}\|_2\right),
\end{align}
where
$\frac{\partial  \mathcal{H}(P_Y)}{\partial P_{Y}(y)}=-\log P_Y(y)-\log(e)$ for each $y\in\mathcal{Y}$. Then using Cauchy-Schwartz inequality we have:
\begin{align}
\left| \mathcal{H}(P_Y)- \mathcal{H}(\hat{P}_Y)\right| &\leq\left|\left\langle\left\{\log(P_Y(y))\right\}_{y\in\mathcal{Y}},\mathbf{P}_{Y}-\mathbf{\hat{P}}_{Y}\right\rangle\right|+o\left(\|\mathbf{P}_{Y}-\mathbf{\hat{P}}_{Y}\|_2\right)\nonumber\\
&\leq \sqrt{\mathbb{V}\big(\{\log P_Y({y})\}_{y\in\mathcal{Y}}}\big)\left\|\mathbf{P}_{Y}-\mathbf{\hat{P}}_{Y}\right\|_2+o\left(\|\mathbf{P}_{Y}-\mathbf{\hat{P}}_{Y}\|_2\right).
\end{align}
Lemmas \ref{lem:diverg} and \ref{lem:pvector} allow us to bound with an arbitrary probability close to one the terms: $\mathcal{D}\left(\hat{P}_{XY}\|P_{XY}\right)$, $\|\mathbf{P}_{X}-\mathbf{\hat{P}}_{X}\|_2$, $\|\mathbf{P}_{Y}-\mathbf{\hat{P}}_{Y}\|_2$, $\|\mathbf{P}_{XY}-\mathbf{\hat{P}}_{XY}\|_2$ and $\| \mathbf{P}_{X|Y}(\cdot|y)-\mathbf{\hat{P}}_{X|Y}(\cdot|y)\|_2$ for $y\in\mathcal{Y}$ simultaneously. Lemma \ref{lem:simultaneidad} guarantees  that the bounds hold simultaneously over all these $|\mathcal{Y}|+4$ quantities, by replacing $\delta$ with $\delta/(|\mathcal{Y}|+4)$. Then, with probability at least $1-\delta$ the following bounds hold:
\begin{align}
\max\big\{\big\|\mathbf{P}_{XY}-&\mathbf{\hat{P}}_{XY}\big\|_2,\big\|\mathbf{P}_{Y}-\mathbf{\hat{P}}_{Y}\big\|_2,\big\|\mathbf{P}_{X}-\mathbf{\hat{P}}_{X}\big\|_2,\big\|\mathbf{P}_{X|Y}(\cdot|y)-\mathbf{\hat{P}}_{X|Y}(\cdot|y)\big\|_2\big\}\nonumber\\
&\leq\frac{1+\sqrt{\log\left(\frac{|\mathcal{Y}|+4}{\delta}\right)}}{\sqrt{n}}\equiv\frac{B_\delta}{\sqrt{n}},\\
\mathcal{D}\left(\hat{P}_{XY}\|P_{XY}\right)&\leq |\mathcal{X}||\mathcal{Y}|\frac{\log(n+1)}{n}+\frac{1}{n}\log\left(\frac{|\mathcal{Y}|+4}{\delta}\right) \mathcal{O}\left(\frac{\log(n)}{n}\right).
\end{align}
Then, using Lemmas~\ref{lem:cotaphi} and~\ref{lem:cotav}, with probability at least $1-\delta$ we have:
\begin{align}
\mathcal{E}_{\textrm{gap}}(Q,\mathcal{S}_n)&\leq\mathcal{O}\left(\frac{\log(n)}{n}\right)+ \frac{B_\delta}{\sqrt{n}}{\Lambda}\big(Q_{U|X},{Q}_{Y|U},Q_{\hat{Y}|U}\big)\nonumber\\
&+2\sum_{u\in\mathcal{U}}\phi\left(\frac{B_\delta}{\sqrt{n}}\sqrt{\mathbb{V}(\{Q_{U|X}(u|x)\}_{x\in\mathcal{X}})}\right)+\frac{B_\delta}{\sqrt{n}}\sqrt{|\mathcal{Y}|}\log|\mathcal{U}|\nonumber\\
&+\sqrt{\mathbb{V}\big( \{\log P_Y({y})\}_{y\in\mathcal{Y}} \big)}\frac{B_\delta}{\sqrt{n}}\\
&\leq\mathcal{O}\left(\frac{\log(n)}{n}\right)+ \frac{B_\delta}{\sqrt{n}}{\Lambda}\big(Q_{U|X},{Q}_{Y|U},Q_{\hat{Y}|U}\big)\nonumber\\
&+\frac{\log(n)}{\sqrt{n}}B_\delta\sum_{u\in\mathcal{U}}\sqrt{\mathbb{V}(\{Q_{U|X}(u|{x})\}_{x\in\mathcal{X}} )} +\frac{2|\mathcal{U}|e^{-1}+B_\delta\sqrt{|\mathcal{Y}|}\log|\mathcal{U}|}{\sqrt{n}}+\sqrt{\mathbb{V}\big(\{\log P_Y({y})\}_{y\in\mathcal{Y}} \big)}\frac{B_\delta}{\sqrt{n}}\nonumber\\
&=\mathcal{O}\left(\frac{\log(n)}{n}\right)+\frac{\log(n)}{\sqrt{n}}A_\delta\sqrt{\mathcal{I}(P_X;Q_{U|X})}+\frac{B_\delta{\Lambda}\big(Q_{U|X},{Q}_{Y|U},Q_{\hat{Y}|U}\big)+C_\delta}{\sqrt{n}}.
\end{align}
Finally, using Taylor arguments as above, we can easily write:
\begin{align}
\left|\sqrt{\mathcal{I}\left(P_X;Q_{U|X}\right)}-\sqrt{\mathcal{I}\left(\hat{P}_X;Q_{U|X}\right)}\right| &\equiv  \mathcal{O}(\|\mathbf{P}_{X}-\mathbf{\hat{P}}_{X}\|_2)\leq \mathcal{O}(n^{-1/2}),\\
\left| {\Lambda}\big(Q_{U|X},{Q}_{Y|U},Q_{\hat{Y}|U}\big) -{\Lambda}\big(Q_{U|X},\hat{Q}_{Y|U},Q_{\hat{Y}|U}\big)\right|&\equiv  \mathcal{O}(\|\mathbf{P}_{XY}-\mathbf{\hat{P}}_{XY}\|_2)\leq \mathcal{O}(n^{-1/2}),
\end{align}
with probability $1-\delta$. 

The last step follows by applying the identity $\mathbb{E}_W\left[ \big(f(W) - \mathbb{E}_T[f(T)] \big)^2 \right] \leq \mathbb{E}_W\mathbb{E}_T \left[ \big(f(W) - f(T) \big)^2 \right]$ to the decoder expression $\Lambda \big(Q_{U|X},{Q}_{Y|U},Q_{\hat{Y}|U}\big)$ and $\mathbb{V}\big(\{\log P_Y({y})\}_{y\in\mathcal{Y}} \big)$. This concludes the proof.
\end{IEEEproof}

\section{Experimental Results}\label{Section-6}

In this section, the interplay between generalization and information complexity of deep neural networks is  empirically studied. This study is indeed to share  light on possible implications of the results in Theorem~ \ref{thm:regularizar} from a practical perspective. As our main goal is not to present a new classification methodology comparing results with state-of-the-art methods, we restrict ourself to use small databases in order to facilitate numerical simulations, as motivated in~\cite{DBLP:journals/corr/NeyshaburTS14}. We train a baseline $3$-layer feed-forward neural network (based on logistic activation) using dropout on two different random subsets of: MNIST (standard data set of handwritten digits) and CIFAR-10 (natural images \cite{cifar}). The size of the training set is $5K$ for both datasets, and early stopping is employed based on the misclassification error according to a random  selection of the validation set composed of $500$ samples. Following suggestions from~\cite{srivastava2014dropout}, the dropout rate $p_{\rm out}$ is confined to the interval $[0.5,0.8]$.  The risk $\mathcal{L}(Q)$ is approximated with the testing dataset while $\mathcal{L}_{\text{emp}}(Q,{\mathcal{S}_n})$ with the training dataset and $\mathcal{I}\big(\hat{P}_X;Q_{U|X}\big)$ with the training dataset but using the upper bound introduced  in Section \ref{subsec:feedforward}, as being a surrogate of the information complexity. Experiments are repeated five times and then averaged. The present concerned only with the following questions: 
\begin{enumerate}
\item[1.] How does the information  complexity behave with the network size (number of hidden units)?
\item[2.] How does the aggregate level of noise (e.g., dropout) impact the information complexity? 
\item[3.] Can the information complexity predict meaningful statistical variations of data, e.g., induced  from a mismatch or non-meaningful statistical origin, during the training phase? 
\end{enumerate}
\begin{figure}[t]
\centering
\includegraphics[width=12cm]{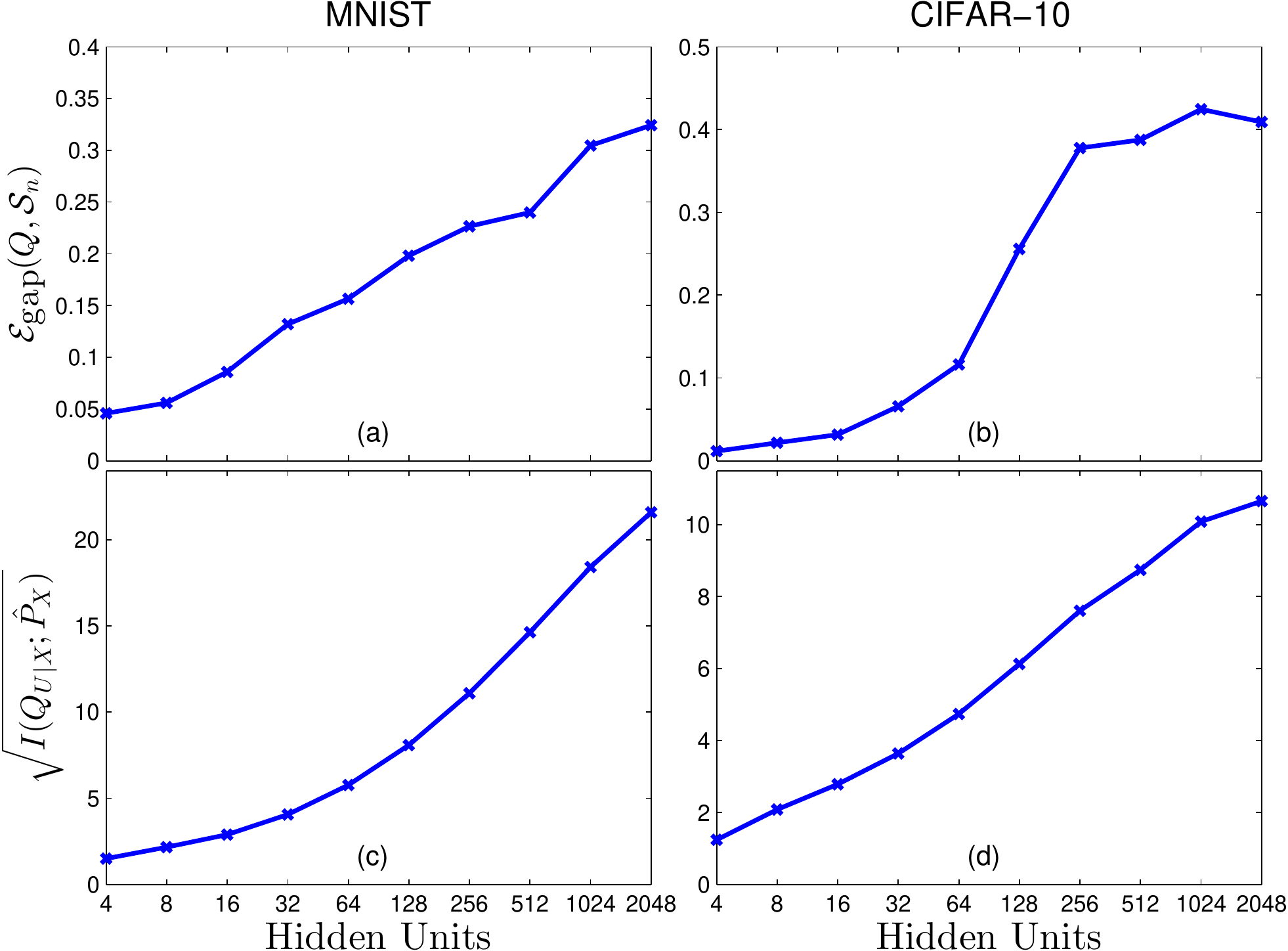} 
\caption{Effect of increasing the number of hidden units: (a) Generalization gap on MNIST dataset and (b) CIFAR-10 dataset, respectively; (c) Information complexity with MNIST and (d) CIFAR-10, respectively.} 
\label{fig:vsm}
\end{figure}

\subsection{Effect of hidden network size on the information complexity}

We begin by studying the interplay between the information complexity  and the network size. In order to compare the behaviour of the information complexity with that of the generalization gap, we run simulations with $3$ layers of feed-forward neural networks with same number of hidden units per each layer and dropout ($p_{\rm out}=0.8$) at each layer. A range of values between $4$ and $2048$ hidden units have been chosen~\cite{DBLP:journals/corr/NeyshaburTS14}. Fig.~\ref{fig:vsm}.(a) and~\ref{fig:vsm}.(b) show the generalization gap as a function of the number of hidden units for MNIST and CIFAR-10, respectively, while Fig.~\ref{fig:vsp}.(c) and \ref{fig:vsp}.(d) show the corresponding surrogate to the information complexity $\sqrt{\mathcal{I}\big(\hat{P}_X;Q_{U|X}\big)}$. First of all, we remark that both curves are consistent with the theoretical expectations and behave similarly in both databases. More precisely, there is an increasing monotonous tendency indicating the presence of possible overfitting when the network size is incremented and  trained in a small dataset. 

\subsection{Effect of noise level on the information complexity}\label{subsec:dropout}

\begin{figure}[t]
\centering
\includegraphics[width=12cm]{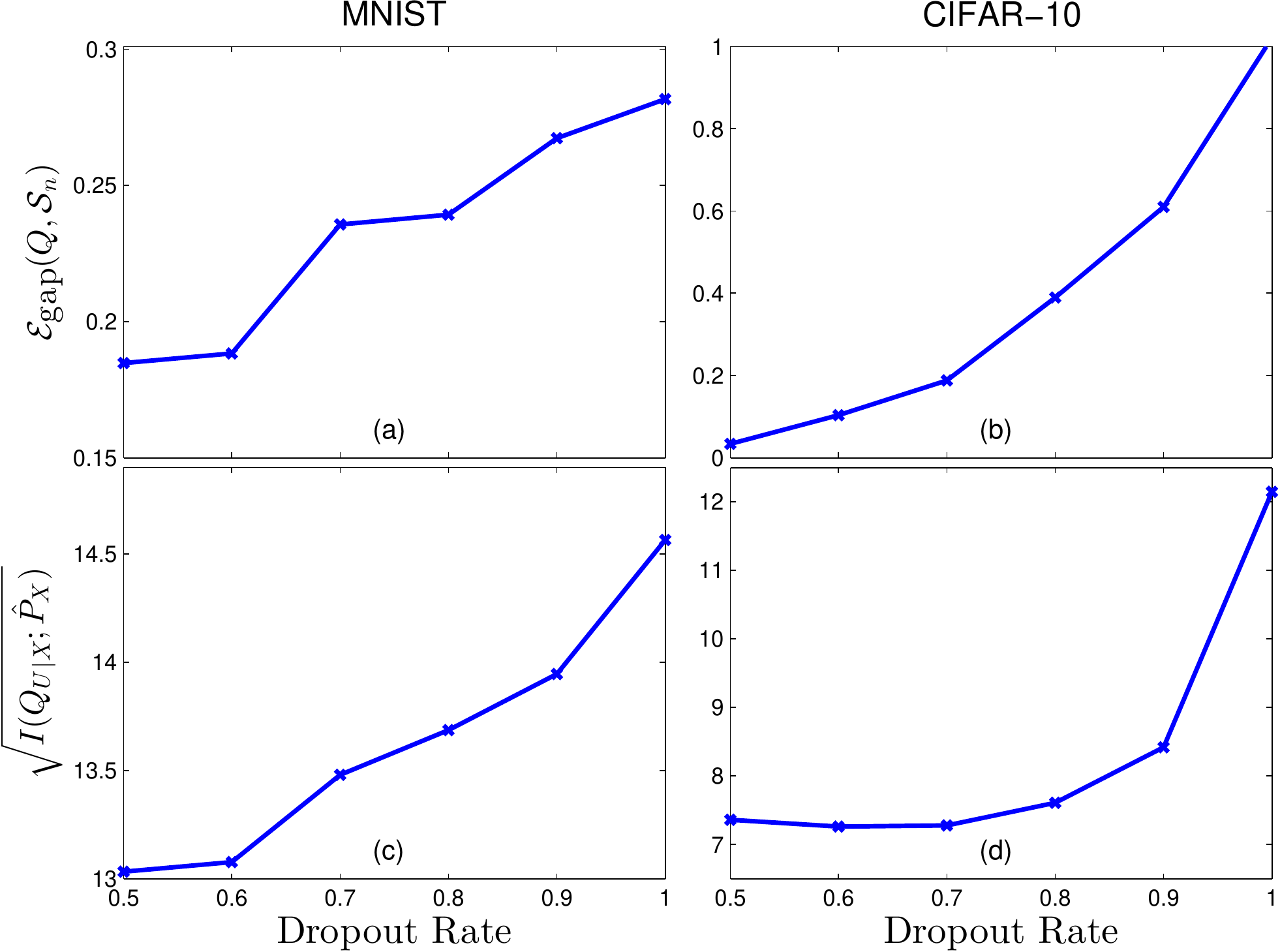} 
\caption{Noise effects on the information complexity: (a) Generalization gap with MNIST dataset and (b) CIFAR-10 dataset, respectively; (c) Information complexity with MNIST and (d) CIFAR-10, respectively.} 
\label{fig:vsp}
\end{figure}
We explore the noise effects on the information complexity for which we request the quantity $m\cdot p_{\rm out}$ to remain constant, where $m$ is the number of hidden units. In this way, we ensure that the expected number of present units will be the same across different (random) architectures. However, the networks for testing will be of different sizes. In our experiments, we set $m\cdot p_{\rm out}=256$  for the first two hidden layers and $m\cdot p_{\rm out}=512$  for  the  last  hidden  layer as was the case in~\cite{srivastava2014dropout}.  Again we notice that the behaviour of both curves is similar for both databases. These figures confirm the expected ability of noise to control generalization, i.e., the network is more regularized as $p_{\rm out}$ goes down. On the other hand, the CIFAR-10 gap described  in Fig.~\ref{fig:vsp}.(b) is more sensitive to $p_{\rm out}$ than MNIST Fig.~\ref{fig:vsp}.(a), this effect can be predicted from the information complexity, as shown in Figs.~\ref{fig:vsp}.(c) and.~\ref{fig:vsp}.(d). Obviously, it seems reasonable that CIFAR-10 where data structure is more richer  requires more regularization than MNIST. 

\subsection{Information complexity can distinguish between different  training sets}

We exemplify the relation between the generalization gap and the information complexity on a mismatch dataset. To this end, we explore the sensitivity of the  information complexity with respect to random and roll labels variation of the training dataset. To this end, we train the network with a subset of the dataset which is used as the initial state of a training session with a new dataset. Similar to the application motivated by \emph{adversarial examples} in \cite{Bartlett17}.
\begin{figure}[t]
\centering
\includegraphics[width=13cm]{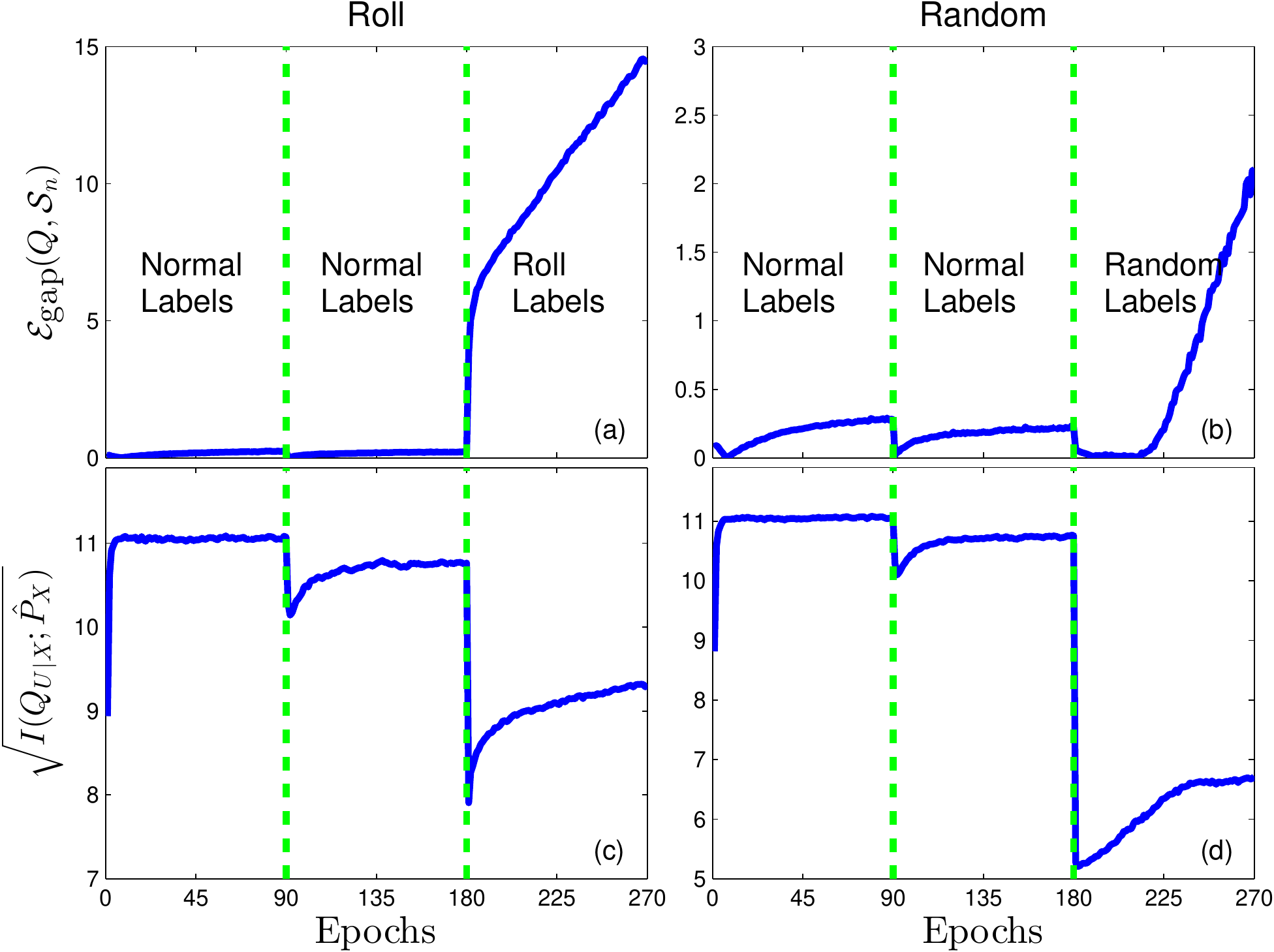} 
\caption{Generalization gap and information complexity on mismatch datasets: (a) Generalization gap for Normal-Normal-Roll labels training set and (c) its information complexity; (b) Generalization gap for Normal-Normal-Random labels training and (d) its information complexity.} 
\label{fig:task}
\end{figure}
The idea is to investigate whether  that information complexity would be able to detect this phase transition between the statistical properties of the training dataset and thus, to distinguish between regular and mismatch labels.  We propose the following experiment: A  feed-forward neural network composed of layers $784$-$256$-$256$-$256$-$10$ with dropout ($p_{\rm out}=0.8$),  trained during $90$ epochs on $5K$ MNIST samples. Then, this network is retrained during $90$ epochs with other $5K$ MNIST training samples. Finally, it is retrained during $90$ epochs with other $5K$ MNIST training samples but with corrupt labels. We consider the cases:
\begin{itemize}
\item {\it Normal labels:} The original labels corresponding to the data; 
\item {\it  Roll labels:} The labels are shifted, i.e., $2\rightarrow 3$, $3\rightarrow 4$ and so on; 
\item {\it Random labels:} All labels are replaced with random ones, according to an uniform distribution.
\end{itemize}
Although the results in Theorem~\ref{thm:regularizar} fails to apply in the present scenario (roll or random labels) since  it assumes that data samples are generated according to the same (true) unknown distribution $P_{XY}$, this does not invalidate the use of the 
information complexity as a mean to predict the generalization ability of the network from a training dataset only. Fig.~\ref{fig:task}.(a) shows the gap for the roll labels case as the training progresses. As expected, the gap increases when a phase transition takes place to the roll labels because the network learns a different (incorrect) task. A significative  difference  can also be observed in terms of the information complexity when this phase transition happens, as shown in Fig. \ref{fig:task}.(c). Indeed, there is a clear  transition in comparison with the normal training labels while also a slower convergence. Similarly, Fig.~\ref{fig:task}.(b) shows the gap for the random labels case. During training from random labels, the network takes time before start learning again, as the difficulty of the task is increased (the generalization increases), as was shown in \cite{DBLP:journals/corr/ZhangBHRV16}. Again, 
the phase transmission can also be detected from the information complexity~Fig.~\ref{fig:task}.(d). Therefore, the information complexity may be used for detecting from training samples only (without validation set) whether a dataset is similar or not to a previous dataset. However, in order to provide mathematical guarantees on this rather complex task, much additional work is required which goes beyond the scope of this paper.

\section{Summary and Further Observations}\label{Section-7}

The results in this paper (Theorem~\ref{thm:regularizar}) show that if the representation model is used for a pattern classification problem by minimizing the cross-entropy loss, the generalization performance depends on the information complexity (IC), i.e., the empirical mutual information between the inputs and their corresponding representations. In this sense, we may speak of IC of representations, meaning the minimum amount of information (measured in bits) we need to store enough information about the representations that allows us its reconstruction. Our main result in this chapter states that good representation models should squeeze out as much regularity as possible from the given data. Compared with the powerful machinery of statistical learning theory, it does not directly involve the covering number or the VC dimension of the overall loss function itself, which in general leads to considerably difficulties for evaluation.

The concept of encoder capacity introduced in Definition~\ref{definition-encoder-capacity} provides a fundamental measure of the generalization capabilities of a encoder model. This observation is naturally driven by Proposition~\ref{lemma-capacity-encoder} that shows an interesting link between the misclassification error of distinguishing input samples from their representations, the encoder capacity and the IC.  As a matter of fact, we can conclude that encoders inducing a large misclassification probability on input samples from their representations are expected to achieve better generalization and to enforce invariant representations. From this perspective, noise injection appears to be a natural way to enforce statistical invariance (e.g. inputs must share similar representations) in representation models. Applications examples were studied for GANs and a restricted RBM where it is shown that there exists a non-trivial solution for the required amount of noise optimizing the surrogate of the risk. We finally presented a simple experimental framework for understanding the role of the IC in the generalization capabilities of standard deep neural networks. The experiments we conducted provide  supporting evidence indicating that there is an apparent link between the generalization ability of deep neural networks and the IC. Though our results might be improved in several directions, the notions of encoder capacity and the role of IC  may be useful for two purposes: to explain why certain architectures generalize better than others and to obtain some hints on the behaviour of the gap.  However, several questions remain open and  further work is needed. 

\appendices

\section{Proof of Lemma~\ref{lemma-surogate-representations}}\label{app:lemma-surogate}
The upper bound simply follows by using Jensen-Inequality~\cite{cover} while the lower bound is a consequence of the definition of the rate-distortion and distortion-rate functions.  
The probability of misclassification  corresponding to the classifier  can be expressed by the expected distortion $ \mathbb{E}_{{P}_{XY} {Q}_{U|X} }\left[d(Y,U) \right]=P_{\mathcal{E}}(Q_{\hat{Y}|U},Q_{{U}|X})$ based on the fidelity function $d(y,u)\coloneqq 1- Q_{\hat{Y}|U}(y|u)$. Because of the Markov chain $Y \mkv X \mkv U$, we can use the data processing inequality \cite{cover} and the definition of the rate-distortion  function, obtaining the following bound for the classification error: 
\begin{IEEEeqnarray}{rCl}\label{eq:boundmi}
\mathcal{I}(P_X;Q_{U|X}) \geq \mathcal{I}(P_Y;Q_{U|Y}) &\geq &\hspace{-0.4cm}\min_{\rule{0mm}{4.3mm}\substack{P_{\hat{U}|Y}\,:\, \mathcal{Y}\rightarrow\mathcal{P}(\mathcal{U}) \\  \mathbb{E}_{P_{\hat{U}Y} }[d(Y,\hat{U})] \, \leq\, \mathbb{E}_{{P}_{XY} {Q}_{U|X} }\left[d(Y,U) \right]}}\hspace{-0.4cm} \mathcal{I}\big(P_Y;P_{\hat{U}|Y}\big)\nonumber\\
&=& \mathcal{R}_{Y,Q_{\hat{Y}|U}}\big(P_{\mathcal{E}}(Q_{\hat{Y}|U},Q_{{U}|X}) \big).\nonumber
\end{IEEEeqnarray}
For $\mathbb{E}_{{P}_{XY} {Q}_{U|X} }\left[d(Y,U) \right]$, 
we can use the definition of $\mathcal{R}_{Y,Q_{\hat{Y}|U}}^{-1}(\cdot)$, and thus obtain from \eqref{eq:boundmi} the fundamental bound $\mathcal{R}_{Y,Q_{\hat{Y}|U}}^{-1}(\mathcal{I}(P_X;Q_{U|X}))\leq \mathcal{R}_{Y,Q_{\hat{Y}|U}}^{-1}(\mathcal{I}(P_Y;Q_{U|Y}))\leq P_{\mathcal{E}}(Q_{\hat{Y}|U},Q_{{U}|X}) $.

\section{Proof of Proposition~\ref{lemma-capacity-encoder}}~\label{app:lemma-capacity-encoder}
We begin with the proof of the lower bound~\eqref{eq-upper-bound-IC}. Consider the following inequalities:
\begin{align}
\mathcal{I}\big(\hat{P}_X;Q_{U|X} \big) &\leq   \min_{Q_U \in\mathcal{P}(\mathcal{U})}     \mathcal{D}\big( Q_{U|X}\big \| Q_U\big|\hat{P}_X \big)  \\
& \leq  \min_{Q_U \in\mathcal{P}(\mathcal{U})}   \mathbb{E}_{\hat{P}_X}  \mathbb{E}_{Q_{U|X}}\left[ \max_{x\in\mathcal{A}_n } \log \frac{Q_{U|X}(U|x)}{Q_U(U)} \right]  \\
& \leq  \min_{Q_U \in\mathcal{P}(\mathcal{U})}  \max_{u \in \mathcal{U}}\,  \log \frac{Q_{U|X}\big(u|\psi^\star ( u)\big)}{Q_U(u)}  \label{eq-regret1}  \\
& =   \log \left(\sum\limits_{u\in\mathcal{U}} Q_{U|X}\big(u|\psi^\star(u)\big) \right)  =  \mathcal{C}_e\big(Q_{U|X},\mathcal{A}_n\big),  \label{eq-regret2}
\end{align}
where inequality~\eqref{eq-regret1} follows by letting $\psi^\star$ to be the mapping maximizing  $ \mathcal{C}_e\big(Q_{U|X},\mathcal{A}_n\big)$ and~\eqref{eq-regret2} follows by noticing that~\eqref{eq-regret1} is the smallest worst-case regret, known as the \emph{minimax regret}, and thus by choosing $Q_U $ to be the normalized maximum-likelihood distribution on the restricted set $\mathcal{A}_n$ the claim is a consequence of remarkable result of Shtarkov~\cite{shtarkov87universal}.

It remains to show the bounds in~\eqref{propo-capacity2}. In order to show the lower bound, we can simply apply Fano's lemma~\cite[Lemma 2.10]{Tsybakov:2008:INE:1522486} from which we can bound from below the error probability~\eqref{eq-capacity-encoder-expression2} based on $\mathcal{A}_n$. As for the upper bound, we have:
\begin{align}
\log| \mathcal{A}_n|  - \mathcal{I}\big(\hat{P}_X;Q_{U|X} \big) &\geq   \mathcal{H} \big(\hat{P}_{X} \big)  - \mathcal{I}\big(\hat{P}_X;Q_{U|X} \big) \label{eq-prop-1} \\
 & = \sum\limits_{u\in\mathcal{U}} \hat{Q}_U(u)  \mathcal{H} \big(\hat{Q}_{X|U} (\cdot|u) \big)  \label{eq-prop-2}\\
&\geq 2 \sum\limits_{u\in\mathcal{U}} \hat{Q}_U(u)  \left(  1-  \max\limits_{x^\prime\in\mathcal{X}}  \hat{Q}_{X|U}(x^\prime |u) \right)    \label{eq-prop-3}\\
& = 2 \varepsilon   \label{eq-prop-4}, 
\end{align}
where \eqref{eq-prop-1} follows from the assumption $\mathcal{A}_n=\textrm{supp}\big(\hat{P}_X\big)$ and the fact that the entropy is maximal over the uniform distribution; \eqref{eq-prop-3} follows by using \cite[eq. (7)]{272494} and \eqref{eq-prop-4} by the definition of $\varepsilon$ in \eqref{eq-capacity-encoder-expression2}. This concludes the proof of the proposition.

\section{Proof of Proposition~\ref{lemma:lower_emp_pe}}\label{app:lemma-lower_bound_pe}
The proof follows from Lemma~\ref{lemma-surogate-representations} and the observation that $R\mapsto \mathcal{R}_{Y,Q_{\hat{Y}|U}}^{-1}(R)$ is a convex function. In fact, consider $I=\lambda I_1+(1-\lambda)I_2$ with $\lambda\in[0,1]$. From the definition of $\mathcal{R}_{Y,Q_{\hat{Y}|U}}^{-1}(\cdot)$ in expression \eqref{eq:inverse_rate} and the decreasing property of $D\mapsto \mathcal{R}_{Y,Q_{\hat{Y}|U}}(D)$, it is clear that $\mathcal{R}_{Y,Q_{\hat{Y}|U}}\left(\mathcal{R}_{Y,Q_{\hat{Y}|U}}^{-1}(I_1)\right)=I_1$ and $\mathcal{R}_{Y,Q_{\hat{Y}|U}}\left(\mathcal{R}_{Y,Q_{\hat{Y}|U}}^{-1}(I_2)\right)=I_2$. As $D\mapsto \mathcal{R}_{Y,Q_{\hat{Y}|U}}(D)$ is convex we have that $\mathcal{R}_{Y,Q_{\hat{Y}|U}}\big(\lambda \mathcal{R}_{Y,Q_{\hat{Y}|U}}^{-1}(I_1)+(1-\lambda)\mathcal{R}_{Y,Q_{\hat{Y}|U}}^{-1}(I_2)\big)\leq \lambda I_1+(1-\lambda)I_2$. From \eqref{eq:inverse_rate} and the decreasing property of $R\mapsto \mathcal{R}_{Y,Q_{\hat{Y}|U}}^{-1}(R)$ we can write:
\begin{align}
\mathcal{R}_{Y,Q_{\hat{Y}|U}}^{-1}(I)&=\inf\left\{D\in\mathbb{R}_{\leq 0}:\mathcal{R}_{Y,Q_{\hat{Y}|U}}(D)\leq\lambda I_1+(1-\lambda)I_2\right\},\nonumber\\
&\leq\lambda \mathcal{R}_{Y,Q_{\hat{Y}|U}}^{-1}(I_1)+(1-\lambda)\mathcal{R}_{Y,Q_{\hat{Y}|U}}^{-1}(I_2),
\end{align}
obtaining the desired result. As $R\mapsto \mathcal{R}_{Y,Q_{\hat{Y}|U}}^{-1}(R)$ is convex, and assuming that it is differentiable at the point of interest $\mathcal{I}(P_Y;Q_{U|Y})$\footnote{Some mild conditions for this to hold can be found in \cite{csiszar_1974}.}, we obtain:
\small
\begin{align}
\mathcal{R}_{Y,Q_{\hat{Y}|U}}^{-1}\big(\mathcal{I}(P_Y;Q_{U|Y})\big)\geq \mathcal{R}_{Y,Q_{\hat{Y}|U}}^{-1}\left( \mathcal{I}(\hat{P}_Y;\hat{Q}_{U|Y})  \right) &+\left[\mathcal{I}(P_Y;Q_{U|Y})-\mathcal{I}(\hat{P}_Y;\hat{Q}_{U|Y})\right]\frac{\partial \mathcal{R}_{Y,Q_{\hat{Y}|U}}^{-1}(I)}{\partial I}\Big|_{I=\mathcal{I}(\hat{P}_Y;\hat{Q}_{U|Y})}\nonumber
\end{align}\normalsize
where $\hat{Q}_{U|Y}(u|y)=\sum_{x\in\mathcal{X}}Q_{U|X}(u|x)\hat{P}_{X|Y}(x|y)$ for $(u,y)\in\mathcal{U}\times\mathcal{Y}$. Observe that this bound holds with probability one. Using the fact that $\mathcal{I}(P_Y;Q_{U|Y})$ is  differentiable with respect to $\mathbf{P}_{XY}$, we have:
\begin{align}
\mathcal{I}(\hat{P}_Y;\hat{Q}_{U|Y})-\mathcal{I}(P_Y;Q_{U|Y})= \left\langle\frac{\partial \mathcal{I}(P_Y;Q_{U|Y})}{\partial \mathbf{P}_{XY}},\mathbf{\hat{P}}_{XY}-\mathbf{P}_{XY}\right\rangle+o\left(\|\mathbf{P}_{XY}-\mathbf{\hat{P}}_{XY}\|_2\right).\nonumber
\end{align}
After some lengthy but standard algebraic calculations, it can be shown that:
\begin{equation}
\frac{\partial \mathcal{I}(P_Y;Q_{U|Y})}{\partial P_{XY}(x',y')}=\sum_{u\in\mathcal{U}}Q_{U|X}(u|x')\log\left(\frac{Q_{U|Y}(u|y')}{Q_{U}(u)}\right)-1,
\label{eq:inf_mutua_deriv}
\end{equation}
for all $(x',y')\in\mathcal{X}\times\mathcal{Y}$. It is immediate to see that:
\begin{align}
\left|\mathcal{I}(\hat{P}_Y;\hat{Q}_{U|Y})-\mathcal{I}(P_Y;Q_{U|Y})\right| \leq \left\|\frac{\partial \mathcal{I}(P_Y;Q_{U|Y})}{\partial \mathbf{P}_{XY}}\right\|_2\|\mathbf{\hat{P}}_{XY}-\mathbf{P}_{XY}\|_2+o\left(\|\mathbf{P}_{XY}-\mathbf{\hat{P}}_{XY}\|_2\right).
\end{align}
From (\ref{eq:inf_mutua_deriv}) and the assumptions of our statistical model we have that:
\begin{equation}
\left\|\frac{\partial \mathcal{I}(P_Y;Q_{U|Y})}{\partial \mathbf{P}_{XY}}\right\|_2 \leq \sqrt{|\mathcal{X}\|\mathcal{Y}|}\left[\log\left(\frac{1}{P_Y(y_{\min})}\right)-1\right].
\end{equation}
The result is finally obtained from Lemma~\ref{lem:pvector} and the fact that $\mathcal{R}_{Y,Q_{\hat{Y}|U}}^{-1}(\cdot)\geq 0$.

\section{Auxiliary Results}\label{app:auxiliary-results}

In this appendix some auxiliary results, which are used in proof of the main results are listed. The majority of them are presented without proof, because are either straightforward or are easily found in the bibliography. Only a few results are presented with their corresponding proofs.

\begin{lemma}[Pinsker's inequality]\label{lem:Dineq}
	For any pair of probability distributions $P_X,Q_X$:
	\begin{equation}
	\frac{1}{2}\mathcal{D}(P_X\|Q_X)\geq\left(\sum_{x\in\mathcal{X}}\big|P_X(x)-Q_X(x)\big|\right)^2.
	\end{equation}
\end{lemma}

\begin{theorem}[McDiarmid's Inequality \cite{Devroye97a}]\label{teo:mcdiarmid}
	Let $\mathcal{S}_n=\{Z_k\}_{k=1}^n$ i.i.d. random variables, and let $f:\mathcal{Z}^n\rightarrow\mathbb{R}$ a bounded difference function i.e.:
	\begin{equation}
	\big|f(z_1,\cdots,z_k,\cdots,z_n)-f(z_1,\cdots,z_k^\prime,\cdots,z_n)\big|\leq C,
	\end{equation}
	for all $\,1\leq k\leq n\;\forall z_1,\cdots,z_n,z_k^\prime\in\mathcal{Z}$. Then, for all $\delta\in(0,1]$
	\begin{align}
	{\mathbb{P}}\left(\big|f-\mathbb{E}[f]\big|\geq C\sqrt{\frac{n\log(2/\delta)}{2}}\right)&\leq\delta,\\
	{\mathbb{P}}\left(f-\mathbb{E}[f]\geq C\sqrt{\frac{n\log(1/\delta)}{2}}\right)&\leq\delta.
	\end{align}
\end{theorem}

\begin{lemma}[Theorem 12.2.1 \cite{cover}]\label{lem:diverg}
Let ${P}\in\mathcal{P}(\mathcal{X})$ be a probability distribution and let $\hat{P}$ be its empirical estimation over a $n$-data set $\mathcal{S}_n$. Then, with a probability of at least $1-\delta$ over $\mathcal{S}_n$, 
\begin{equation}
\mathcal{D}(\hat{P}\|P)\leq |\mathcal{X} | \frac{\log(n+1)}{n}+\frac{1}{n}\log(1/\delta).
\end{equation}
\end{lemma}


\begin{lemma}\label{lem:simultaneidad}
Let $\{\mathcal{A}_k\}_{k=1}^m$ events such that $\Pr(\mathcal{A}_k)\geq1-\delta$ for each $k\in[1: m]$. Then
\begin{equation}\label{eq:induction_simultaneos}
\mathbb{P}\left(\bigcap_{k=1}^m \mathcal{A}_k\right)\geq1-\delta m.
\end{equation}
\end{lemma}

\begin{lemma}\label{lem:pvector}
Let ${P}\in\mathcal{P}(\mathcal{X})$ be any probability distribution and let $\hat{{P}}$ be its empirical estimation over a $n$-data set $\mathcal{S}_n$. Then, with probability at least $1-\delta$ over $\mathcal{S}_n$, the following inequality holds:
\begin{equation}
\|\mathbf{P}-\mathbf{\hat{{P}}}\|_2\leq\frac{1+\sqrt{\log(1/\delta)}}{\sqrt{n}}.
\end{equation}
\end{lemma}
\begin{IEEEproof}
The proof is an straightforward application of Theorem \ref{teo:mcdiarmid} with $f\equiv\|\mathbf{P}-\mathbf{\hat{{P}}}\|_2$ jointly with a symmetrization argument (see \cite{Devroye97a}) for tight bounding of $\mathbb{E}\|\mathbf{P}-\mathbf{\hat{{P}}}\|_2$. For these reasons is omitted.
\end{IEEEproof}

\begin{lemma}\label{lem:cotaphi}
Let $n\geq a^{2}e^{2}$, then
\begin{equation}
\phi\left(\frac{a}{\sqrt{n}}\right)\leq\frac{a}{2}\frac{\log(n)}{\sqrt{n}}+\frac{e^{-1}}{\sqrt{n}},
\end{equation}
where
\begin{equation}
\phi(x)=\left\{\begin{array}{cc}0&x\leq0\\-x\log(x)&0<x<e^{-1}\\e^{-1}&x\geq e^{-1}
\end{array}\right.
\label{eq:cotaphi}
\end{equation}
\end{lemma}
\begin{IEEEproof}
Note that $-x\log(x)\leq e^{-1}$ for all $x\in[0,e^{-1}]$. If $n$ is large enough, $n\geq a^2e^2$, then
\begin{equation}
\phi\left(\frac{a}{\sqrt{n}}\right)=\frac{a}{2}\frac{\log(n)}{\sqrt{n}}-\frac{a\log(a)}{\sqrt{n}}\leq\frac{a}{2}\frac{\log(n)}{\sqrt{n}}+\frac{e^{-1}}{\sqrt{n}}.
\end{equation}
\end{IEEEproof}


The following two lemmas will play an important role in the proof of the main results. These will allow us to obtain a bound on
$\left| \mathcal{H}(Q_U)- \mathcal{H}(\hat{Q}_U)\right|$ and $\left| \mathcal{H}(Q_{U|Y}|P_Y)- \mathcal{H}(\hat{Q}_{U|Y}|\hat{P}_Y)\right|$ in terms of the encoder $Q_{U|X}$. Their proofs can be found in~\cite{Shamir:2010:LGI:1808343.1808503}.

\begin{lemma}\label{lem:Hu}
	Consider the encoder given by $Q_{U|X}$. We have
	\begin{align}
	\left| \mathcal{H}(Q_U)- \mathcal{H}(\hat{Q}_U)\right|\leq \sum_{u\in\mathcal{U}}\phi\left(\|\mathbf{P}_X-\mathbf{\hat{P}}_X\|_2\cdot\sqrt{\mathbb{V}\big(\{ Q_{U|X}(u|{x})\}_{x\in\mathcal{X}}\big) }\right),
	\end{align}
	where function $\mathbb{V}(\cdot)$ is defined in (\ref{eq:emp_var}).
\end{lemma}
	

\begin{lemma}\label{lem:Hu|y}
	Consider the encoder given by $Q_{U|X}$	
\begin{align}
\left| \mathcal{H}(Q_{U|Y}|P_Y)- \mathcal{H}(\hat{Q}_{U|Y}|\hat{P}_Y)\right|& \leq   \|\mathbf{P}_Y-\mathbf{\hat{P}}_Y\|_2\sqrt{|\mathcal{Y}|}\log|\mathcal{U}|\nonumber\\
+ &\mathbb{E}_{P_Y}\left[ \sum_{u\in\mathcal{U}}\phi\left(\left\|\mathbf{P}_{X|Y}(\cdot|Y)-\mathbf{\hat{P}}_{X|Y}(\cdot|Y)\right\|_2\right. \left.\times\sqrt{\mathbb{V}\big(\{ Q_{U|X}(u|{x})\}_{x\in\mathcal{X}}\big) }\right)\right],
\end{align}
	where function $\mathbb{V}(\cdot)$ is defined in (\ref{eq:emp_var}).
\end{lemma}

The following lemma plays an important role since it allows us to connect the mutual information $\mathcal{I}(P_X;Q_{U|X})$ with the term $\sum_{u\in\mathcal{U}}\sqrt{\mathbb{V}\big(\{ Q_{U|X}(u|{x})\}_{x\in\mathcal{X}}\big) }$. Its proof follows from an application of Pinsker's inequality to Lemma~\ref{lem:Dineq} and can be found in~\cite{Shamir:2010:LGI:1808343.1808503}.
\begin{lemma}\label{lem:cotav}
The following inequality holds:
\begin{align}
\sum_{u\in\mathcal{U}}\sqrt{\mathbb{V}\big(\{ Q_{U|X}(u|{x})\}_{x\in\mathcal{X}}\big) } \leq\frac{\sqrt{2}}{p_{X}(x_{\min})}\left(1+\sqrt{\frac{1}{|\mathcal{X}|}}\right)\sqrt{\mathcal{I}(P_X;Q_{U|X})}.
\end{align}
\end{lemma}

\bibliographystyle{IEEEtran}

%







\bibliography{proposal}

\end{document}